\documentclass[11pt,a4paper]{article}
\usepackage{times,latexsym}
\usepackage{url}
\usepackage[T1]{fontenc}
\usepackage{xcolor}
\usepackage{multirow}
\usepackage{hhline}
\usepackage{vcell}
\usepackage[acceptedWithA]{tacl2021v1}
\usepackage{xspace,mfirstuc,tabulary}

\usepackage{latexsym}
\usepackage{soul}
\usepackage{microtype}
\usepackage{inconsolata}
\usepackage{multirow}
\usepackage{graphicx}
\usepackage{tabularray}
\usepackage{subcaption}
\usepackage{hhline}
\usepackage{arydshln}
\usepackage{siunitx}
\usepackage{newtxmath}

\newcommand{\cutout}[1]{\textcolor{gray}{}}

\newif\iftaclinstructions
\taclinstructionsfalse % AUTHORS: do NOT set this to true
\iftaclinstructions
\renewcommand{\confidential}{}
\renewcommand{\anonsubtext}{(No author info supplied here, for consistency with
TACL-submission anonymization requirements)}
\newcommand{\instr}
\fi

\iftaclpubformat % this "if" is set by the choice of options

\else

\fi

\definecolor{update}{rgb}{0,0,0}
\title{Automated Essay Scoring and Language Certification:\\Assessing Generalizability, Agreement and Validity for French}

\author{
  Rodrigo Wilkens$^\diamondsuit$ %\Thanks{The {\em actual} contributors to this instruction     document and corresponding template file are given in Section     \ref{sec:contributors}.} 
  \and
  Rémi Cardon$^\spadesuit$
  \and
  Vincent Folny$^\heartsuit$
  \and
  Thomas François$^\spadesuit$
  \\
  \ \\
  $^\diamondsuit$ University of Exeter, 
  $^\heartsuit$ France Éducation international, $^\spadesuit$ Cental, IL\&C, UCLouvain
  \ \\ 
  $^\spadesuit$ Computer Science and Engineering Department, Universidad Carlos III de Madrid
  \ \\
  \\
  \texttt{r.wilkens@exeter.ac.uk, rcardon@inf.uc3m.es} \\
  \texttt{Folny@france-education-international.fr thomas.francois@uclouvain.be}
  \\
  \texttt{}
}

\begin{document}
\maketitle

% dire qu'on a 2 contributions

\begin{abstract}
In Automated Essay Scoring (AES), benchmarking practices have fostered minimalist evaluation practices, in contrast with the broader-view recommendations of evaluation frameworks, such as the argument-based validation framework (ABV), which argued in favor of a multidimensional assessment of systems, especially in the context of high-stakes language tests. 
In this paper, we introduce an enhanced and more practical version of the ABV framework, incorporating fairness analysis, correlations with linguistic features, prediction error evaluation, and model agreement compared with human raters. % to provide a more rigorous assessment. % of AES validity. 
Applying this framework to French AES, we compare 8 model architectures %, including CamemBERT and hybrid architectures, 
on a corpus of 27k exam essays (2 raters each) and a generalization corpus of 961 essays (at least nine raters each). Our analyses illustrate the benefits of applying the ABV framework to better understand the capabilities and pitfalls of AES models, while also advancing the state-of-the-art for French AES.
%Automated Essay Scoring (AES) offers methods for assessing language proficiency. While the focus is on researching new models, evaluation practices have barely evolved. We address this issue, aiming to bridge the gap between current evaluation practices and the rigorous requirements of high-stakes language tests.
%}
%Our objective is to assess the validity of AES models by examining raters' agreement and model generalizability, in the context of written French language certification. We present a systematic evaluation exploring fairness assessments, correlations with linguistic features, prediction error analysis, and standard AES metrics. Additionally, we compare models' performance with human raters.
%
%Furthermore, we enhance the assessment framework by integrating quantitative metrics, thereby advancing evaluation practices in the field and identifying potential pitfalls of AES for language certification. Our work also advances the state of the art for French AES.
% \textbf{keyword} AES, French, language certification, model reliability, AES architecture
\end{abstract}
%\section*{Répartition des tâches (en cours)}
%\subsection*{Rémi}
%État de l'art sur les tailles de corpus utilisés en deep learning pour l'AES, et autant que possible l'influence de cette caractéristique sur les performances.
\section{Introduction}

Automated Essay Scoring (AES)\footnote{
Automated Essay Scoring is also called Automated Essay Grading (AEG) \cite{valenti_overview_2003}, Automated Essay Evaluation (AEE) \cite{shermis_contrasting_2015}, Automated Writing Evaluation (AWE) \cite{beigman-klebanov-madnani-2020-automated}, and 
Analytic Writing Assessment (AWA) \cite{rudner_evaluation_2006}.}
aims to produce systems that imitate groups of expert raters by combining principles of educational assessment and Natural Language Processing (NLP). The goal is to provide a valid, cost-efficient and reliable solution for the automatic evaluation of essays \cite{klebanovandmadnani2021}. 
\textcolor{update}{
In this paper, we focus on certification tests\footnote{
For a comprehensive list of certification language tests see \url{en.wikipedia.org/wiki/List_of_language_
proficiency_tests}.}, which are a type of standardized language test defined as ``a process by which individuals are certified as having demonstrated some level of knowledge and skill in some domain'' \cite[p. 216]{aera2014standards}. Certification tests are generally high-stake, meaning that outcomes such as ``admission, promotion, placement or graduation are directly dependent on test scores'' \cite[p. 300]{shohamy1996test} in contrast to low-stake exams (e.g., formative evaluation). In view of the social consequences of high-stake tests, their validity and reliability are critical, as already pointed out by \citet[p. 17]{messick1990}. 
}
% as they impact high-stake decision-making processes such as naturalisation and %residency, 
%access to employment or studies.  

AES systems tend to be scalable and flexible \cite{klebanovandmadnani2021}, and reliable and reproducible \cite{yan2020handbook}, which are desirable characteristics for real-world utilization.
%
%In contrast, AES systems' validity\footnote{Validity is ``the degree to which evidence and theory support the implementation of scores assigned to a test (whether by humans or by automated systems) for proposed uses of tests'' \cite{american2014american}.} is attainable, but not assured by design. Therefore, the question in AES is now %not if we can reach validity
% how validity should be reached 
% and monitored through evaluation \cite{williamson2012framework}. 
%
%However, 
Recent AES literature, especially in NLP, has focused more on models' improvements or new architectures than on the assessment of validity. For ease of comparison, most studies operate within a controlled setting (i.e., %a limited set of prompts, 
the ASAP dataset and a single metric) \cite{kusuma2022automated}.
%This has led to evaluation standards that do not meet the certification needs. 
However, this does not align with the requirements of high-stakes language tests. 
Commonly overlooked elements in AES evaluation include the results' generalizability across different corpora, and the relationship between the model and the human raters' agreement. Yet, those elements are required for standardized language certification tests, such as IELTS or TCF. %\footnote{For a comprehensive list of language certification exams see \url{en.wikipedia.org/wiki/List_of_language_proficiency_tests}.}. 
%There is also a need for %an extensive
%the diversification of metrics for evaluating the validity of a model in relation to the construct \cite{klebanovandmadnani2021}.
% In consequence, there exists a gap between the scientific literature %on AES models and the requirements from language certification exams, such as IELTS, EF Standard English Test, or TCF.

% - Test de connaissance du français
%TELC.%  - The European Language Certificates

%Filling this gap would require expanding the focus to a greater variety of corpora and languages and widening the array of exploited metrics. % and their relation with assessment frameworks (e.g., \cite{}). %for evaluating the validity of a model to model the construct. \cite{klebanovandmadnani2021}. 
% In this paper, we are moving forward in this direction by reporting 

In language testing, evaluating the tests' validity and reliability has been extensively investigated, using the Argument-Based Validation (ABV) framework \cite{chapelle2008building}, best developed by \citet{williamson2012framework}. This tradition has had little impact on NLP practices. Aiming to bridge this gap, we perform
extensive experiments in the context of one of the major standardized language certification tests for French, the French Knowledge Test (\textit{Test de connaissance du Français} -- TCF), and report them within the ABV framework. 

\textcolor{update}{
%As a first contribution, 
This work advances the state-of-the-art for AES, targeting the French language. We compare 8 different model architectures, including hybrid architectures\footnote{\url{https://gitlab.com/rswilkens/linguistic-features-in-transformers}}, relying on the largest corpus made for French AES.
In addition, based on the ABV framework, we report the most extensive evaluation of AES models for French to date, with a strong view on the real-world utilization of our system. We go further in our systematic comparison of various AES approaches, by leveraging a high-quality test corpus made for this specific purpose. % ajouter des références aux sections ?
}

\textcolor{update}{
Beyond this specific contribution for French, our main contribution consists in an enhancement of the ABV framework. As it is not easily automatable in its original form, we have identified relevant automatic metrics from NLP and systematically linked them to the framework. Thus, we offer a practical AES evaluation framework for language testing. The focus of our approach is aligned with the recommendations for adding automated scoring to an existing assessment \cite{madnani-cahill-2018-automated}. We argue that, by bridging the gap between methodologies focused on model validity and more theory-grounded arguments, such a methodological approach will enable enhancements to automatic evaluation practices for AES.
}%In addition, 

The outline of the paper is as follows. We first provide an overview of the literature on AES, focusing on models, their limitations for certified exams, and an introduction of the ABV framework (Section \ref{sec:aesmodels}). We then describe how we approach this framework that will guide our study in Section \ref{sec:Methodology}. Next, we describe the models' architectures we compare (Section \ref{sec:models}) and the corpora we work with (Section \ref{sec:corpus}). Finally, we present the results of our experiments, structured in relation to our enhanced ABV framework (Section \ref{sec:expe}), discuss our findings (Section \ref{sec:disc}), and conclude (Section \ref{sec:ccl}).

%Section \ref{sec:Methodology}

\section{Approaches for AES}
\label{sec:aesmodels}

% \paragraph{AES}
\subsection{AES Models}
\label{aes:models}
AES has a long tradition of feature engineering methods, as they facilitate studying the relation to the test construct. According to \citet{zesch2015task}, features can be classified according to their degree of task dependence, ranging from weak to strong.
% features can be categorized based on their dependency to the task (from weakly to strongly dependent). 
They showed that task-independent features transfer better between prompts, while the task-dependent ones do not generalize well but perform better.
This is especially important in the context of certified exams, where prompts need to be regularly updated.
Examples of feature-based works are \citet{burstein1999automated} with discourse features, %\citet{mahana2012automated} who tested essay length and various punctuation's markers, %\citet{chen2013automated} who also considered syntactic features and important topics, 
or
\citet{zesch2015task}, which considers readability features, grammar errors, and task and topical similarity.

%\citet{ostling2013automated} used essay length, the relative ratio of POS-tags (to detect style preferences of writers);

% essay length \cite{mahana2012automated, chen2013automated},
 %the occurrence of commas, quotations, or exclamation marks \cite{mahana2012automated},
 %syntactic features \cite{chen2013automated},
% readability features \cite{zesch2015task},
% important words or topics \cite{chen2013automated},
% essay length \cite{ostling2013automated},
 %usage of connectors \cite{burstein1999automated},
 %occurrence of causal- and temporal clauses \cite{burstein1998automated, chen2013automated},
% the relative ratio of POS-tags (to detect style preferences of writers) \cite{ostling2013automated},
 %type-token-ratio \cite{zesch2015task},
 %word frequency \cite{zesch2015task},
 %Rhetorical Structure Theory \cite{burstein2001automated},
 %topical overlap ( n-gram overlap) between adjacent sentences \cite{zesch2015task},
 %grammar error \cite{zesch2015task},
% task similarity \cite{zesch2015task}. % (Kullback–Leibler divergence between source and essay) 

AES research then transitioned to deep learning, inspired by the results achieved in other NLP tasks. This led to the development of models independent of linguistic features. \citet{taghipour2016neural} explored CNN and LSTM architectures, replacing linguistic features with word embeddings. They observed that combining results from models trained with different samples and architectures improves performance. 
Building on this, subsequent studies incorporated task-specific knowledge into the models, e.g. coherence as part of the architecture \cite{tay2018skipflow} and rating criteria \cite{liang2018automated}.
% \citet{tay2018skipflow} incorporated coherence information into their SkipFlow architecture by combining word embeddings through a fully-connected layer pair of LSTMs associated with different input tokens. 
% \citet{liang2018automated} proposed SBLSTMA, in which they added information on rating criteria into the model. They used a siamese bi-LSTM architecture where one model receives the essay and the other one receives the rating criteria. While they obtain promising results, reusing their approach on new data is difficult, as the rating criteria were produced manually by domain experts.
% In parallel, 
Yet, \citet{jin2018tdnn} obtained good results by bringing back linguistic features into the models, opening the avenue for hybrid models. 
%To achieve this, they explicitly combined linguistic features (semantic meaning, POS, and syntactic information) with word embeddings.}

Motivated by the strong performance of large language models, \citet{rodriguez2019language,mayfield2020should} explored the use of BERT for AES, obtaining discouraging results.
%Inspired by the good results presented by large language models, \citet{mayfield2020should} explored the performance of BERT for AES. They identified that BERT may produce good results, but at a significant computational cost and, in some cases, with results close to what a simple n-gram model achieves. Despite those discouraging initial results, researchers kept on exploring the task using transformer models.  
\citet{yang2020enhancing} achieved promising results by incorporating multiple loss functions (regression and ranking) in their R2BERT model. 
%\sout{A side effect of the transformer models is the limited input size, which may be critical for AES due to inconsistent performance within the essay. Aiming to fill this gap, \citet{beseiso2021novel} proposed using Bi-LSTM on the top of RoBERTa model to preserve extra recurrent structures.}
Revisiting hybrid methods, \citet{uto2020neural} proposed to concatenate 25 handcrafted essay-level features (length-based, syntactic, word-based, and readability features) to BERT distributed representations.
Exploring generative models, \citet{mizumoto2023exploring} applied  GPT-3 to AES, finding a 52\% exact agreement with the gold-standard, and demonstrating that the inclusion of linguistic features improves GPT-3's predictions.
Finally, \citet{kusuma2022automated}, in their %conducted a systematic 
literature review, found that the best-performing models  -- reaching 0.801 Quadratic Weighted Kappa (QWK) on the ASAP dataset -- are SBLSTMA \cite{liang2018automated} and BERT with handcrafted-features \cite{uto2020neural}. \citet{xie-etal-2022-automated} presented a method combining regression and ranking that obtained 0.817 QWK on ASAP.

% \subsubsection{AES for French}
% \paragraph{AES for French}
As regards AES for French, research remains very limited.
% \cutout{As regards AES for French,
% no sufficiently large corpus is available, making the situation far from encouraging.} 
Early studies employed unsupervised methods: \citet{lemaire_system_2001} used Latent Semantic Analysis (LSA) to compare native students' essays with reference textbooks, whereas AUTO-EVAL \cite{zaghouani_auto-eval_2002} extracted and combined features of L1 essays. Later, \citet{parslow_automated_2015} trained a Naive Bayes classifier on a very small corpus of 200 FFL (French as a Foreign Language) essays. \citet{rankovic_automated_2020} became the first to fine-tune BERT on French data, % on a larger dataset, but did not release the data and the model. % and the dataset only includes one mother tongue.
whereas \citet{wilkens-etal-2023-tcfle} recently released the largest FFL corpus (6.5k essays) and fine-tuned CamemBERT on it. \citet{sanchez-etal-2024-jingle} used the same model and corpus, but focused on freezing layers instead of full fine-tuning. Both works report similar F1 scores (0.56 vs 0.57, with a standard deviation of 0.01 in both approaches). None of these studies has sought to go beyond a performance-driven approach to evaluate their models from a real-world perspective.  

\subsection{Limits for Certification Exams}
\label{sota:limits}
% Some of the standard practices in AES literature are thought to improve the performance and/or the standardization of the models/systems. However, they end up creating practices that limit the application of AES for language proficiency certification tests.

% \textcolor{blue}{falar sobre systemas que dependem de L1}

% However, 
The need to compare models on reference data has led the majority of AES studies to rely on the same English dataset: ASAP %for English 
\citep{kusuma2022automated}. %\st{ASAP essays were written by U.S. native English speakers between grades 7 and 10, so its properties are quite different from those of language certification tests.} %, ASAP \textbf{REF}.
Only a few studies have departed from this perspective, relying on certification exam datasets: for English, \textit{Cambridge Learner Corpus} \citep{nicholls2003cambridge}, the \textit{ETS Corpus} (TOEFL11) \citep{blanchard2013toefl11}, with 12,100 essays written by TOEFL candidates, %of 11 non-English native languages; inspired several AES systems for German, Italian and Czech \citep{vajjala2018experiments,caines2020reprolang,rama2021pre}. 
and \textit{OpenCLC} \cite{open-clc}, which contains 1,238 texts following with the CEFR. % composed of more than 10,000 texts %from candidates of 7 different L1s. \textit{First Certificate of English} (CLC-FCE), contains 1,238 texts aligned with the CEFR \citep{yannakoudakis_new_2011, vajjala_experiments_2018}. 
    %\textit{ETS corpus of non-native written English} or 
    %\textit{TOEFL11} \citep{blanchard2013toefl11}
    %\footnote{The biggest corpus available for certificate purposes with 12,100 English essays written by TOEFL candidates of 11 non-English native languages.} 
For other languages, \textit{MERLIN} \citep{boyd_merlin_2014} with 2,290 texts % contains 2,290 written productions from standardized tests targeting German, Italian (TELC institute) and Czech (UJOP Institute). 
    targeted German, Italian and Czech, 
% Portuguese
    whereas \textit{COPLE2} \citep{mendes-etal-2016-cople2} and %, containing 966 essays written in Portuguese (ICLP and CAPLE institutes), and 
    \textit{ASK} \citep{tenfjord-etal-2006-ask} %, with 1,936 texts written by candidates to the Norwegian Language Test, have also both been used for AES \citep{del_rio_2016, berggren_regression_2019, carlsen_proficiency_2012}. 
    are for Portuguese with 966 and 1936 texts, respectively.
%and 
 %   \textit{TCFLE-8} \cite{wilkens-etal-2023-tcfle} for French. %Note that TCFLE-8 corpus shares its origin with our corpus (i.e., FEI), but the corpus is different.}
Finally, corpora for French are scarce and suffer from shortcomings: they are either too small \cite{parslow_automated_2015} or not available \cite{rankovic_automated_2020}, with the exception of TCFLE-8 \cite{wilkens-etal-2023-tcfle}. 

% with (6k essays).
%large corpus (6k essays) for French (i.e., TCFLE-8 \cite{wilkens-etal-2023-tcfle}) was made available.   
% encouraging, with only a limited number of AES models \cite{lemaire_system_2001, zaghouani_auto-eval_2002, parslow_automated_2015, rankovic_automated_2020}.
% \textbf{ADD TCFLE-8 ; small corpus for training models or simple models as in TCFLE-8.}

Another limitation in AES studies, particularly sensitive for high-stakes language exams, is their exclusive reliance on the QWK metric for evaluation. \citet{doewes2023evaluating} highlighted shortcomings in QWK, emphasizing that relying solely on it could compromise reliability. 
Moreover, most AES models adopt an essay prompt-based approach, requiring one model per essay prompt, which is impractical for language testing\footnote{\citet{jin2018tdnn} identified a performance drop of 0.225 points of QWK %in \citet{taghipour2016neural}’s model  performance 
when comparing prompt-agnostic and prompt-dependent scenarios.}, %due to the fact that
as prompts are frequently updated \cite{jiang-etal-2023-improving}. % to ensure test reliability. 

\subsection{Framework for High-Stakes Assessments}

%\textcolor{red}{
%"Several of these systems are being used operationally to
%score high-stakes assessments in addition to being used as
%the engine for writing practice systems in educational settings. 
%The first among these was \textbf{e-rater} (Burstein, Kukich,
%Wolff, Lu, \& Chodorow, 1998a), which went operational for
%the GMATR in 1999, with the GMAT program later transitioning to IntelliMetric (Rudner et al., 2006) as part of a shift
%of GMAT to a new vendor. The e-rater scoring system is also
%used operationally in conjunction with human scoring for the
%GRE Issue and Argument tasks (Bridgeman, Trapani, \& Attali,
%2009) since October of 2008, for the TOEFL Independent task
%since July of 2009 (Attali, 2009), and for the TOEFL Integrated
%task since November of 2010. Similarly, the IEA automated
%scoring engine is deployed in the Pearson Test of English
%used for high-stakes purposes. However, unlike e-rater, the
%IEA engine is used as the sole rater (Pearson, 2009)."
%}

Some AES implementations are specifically designed for high-stakes assessments. The first system is e-rater \cite{burstein-etal-1998-automated-scoring}, developed for the GMAT test, which uses different feature extraction techniques. This system and others that followed -- like IntelliMetric \cite{rudner_evaluation_2006} -- are exploited conjointly with human raters for tasks within tests such as TOEFL \cite{attali2009evaluating} or GRE \cite{bridgeman2009considering}. One example of an AES system operated without human raters' intervention is IEA for the Pearson Test of English \cite{pearson-pte}, which relies on LSA.

%\textcolor{red}{\textbf{Zhi et al. (2020)} Using Corpus Analyses to Help Address the DIF Interpretation: Gender Differences in Standardized Writing Assessment}

%\subsection{\st{Evaluation of High-Stakes AES}}

High-stakes AES requires specific attention to evaluation \cite{WEIGLE201385, etsreport}. Some researchers, concerned with the validity of AES in high-stakes contexts, have contributed to the development of several validity frameworks for automated scoring \cite{bennettandbejar1998,clauseretal2002,xi2008}. Gradually, these reflections moved closer to Kane's argument-based approach to test validation \cite{kane2013validating}, which is rooted in the field of educational measurement. This trend is illustrated by the works of \citet{clauseretal2002} \citet{xi2008} and \citet{williamson2012framework}, and offers a more comprehensive validation process, encompassing varied evidence sources. More recently, the ABV framework was used by \citet{huawei2023systematic}  as a means of organizing a systematic review of the field. They identified mixed results when comparing different studies, most of which focused on only a few areas of the framework.     

% several scholars have   A widely employed theoretical framework to that end relies on argument-based validation \cite{huawei2023systematic}. This theoretical framework has two main stages \cite{kane2013validating}: the formative stage, related to the design and purpose of a test, and the summative stage, made for empirically checking that the actual implementation complies with the formative test. We focus on the summative stage, as we do not focus on the test itself. 

In this work, we draw from one of the most refined presentations of the ABV framework by 
\citet{williamson2012framework}, who proposed an operationalization of the summative stage \cite{kane2013validating} through five areas of emphasis, which they apply to e-rater for illustration. The strength of \citet{williamson2012framework}'s framework is that it provides several high-level guidelines criteria, and best practices, for each area. In the rest of this section, we briefly introduce these five areas. %, with a focus on what is presented as relevant for fairness analysis.
%detail the guidelines and criteria that are presented as relevant for fairness analysis.

\textit{Construct relevance and representation} (1) focuses on AES model \textit{explanation} by assessing the adequacy between the goals of the assessment (e.g., evaluating linguistic proficiency) and the automated scoring capabilities of the system. 
%This assessment area primarily targets the test 
%-- and in that sense is part of the formative stage --
%but it also states a relation between the scoring and the characteristics of both the construct and the scoring rubrics.
\citet{williamson2012framework} propose four steps to evaluate this fit:
construct evaluation, % (what is the match between the intended construct and the automated scoring capability?),
task design, % (what is the ﬁt between the test task and the features that can be addressed with automated scoring?),
scoring rubric, and % (are the features extracted by the automated scoring mechanism consistent with the features in the scoring rubric?), and
reporting goals. % (are the reporting goals consistent with the automated scoring capability?).

\textit{Association between typical scoring method and human scores} (2) focuses on the \textit{evaluation} and addresses the limitations of using human scores as the ``gold standard'' when evaluating AES systems with automatic metrics (e.g., QWK). The authors suggest a set of additional analyses along with some reference thresholds: first, assess the quality of the human scores; then, report the QWK between human and automated scores using a threshold at which half of the variance of human scores is accounted for by the system (0.70 in their case); measure the difference between system-human and human-human agreements (computed with QWK or correlation) and evaluate that difference against a minimum threshold (-0.10 in their case); report the standardized mean score difference between human and automated scores (standardized on the distribution of human scores), with a maximum threshold of 0.15; assess the threshold required to initiate human adjudication (i.e. using an extra rater); analyze the type of response characteristics that invalidate automatic scoring; and finally, assess the impact of the system on individual tasks or aggregated reported scores by simulating substitutions between automated and human scoring.

\textit{Association with independent measures} (3)
suggests leveraging independent measures, such as scores from another section of the test or external measures of the construct (i.e., \textit{extrapolation}). % to investigate the relationship between human scores and automated scores . 
Specifically, automated scores should have the same relation as human scores on other sections of the test, and align with external measures of the same construct in the same way as human scores. No specific metric is recommended, but the authors acknowledge that divergences hardly occur in AES. 
% , and demonstrate consistent relationships with the three TCF task types and reported CEFR levels.
%This extrapolation could be using the results from other tests, such as speech or reception proficiency. 

%This area aims to evaluate automated scoring in comparison to human scoring, which is an important criterion. %If human and automated scores reflect similar constructs, they are expected to correlate with other measures of similar or distinct constructs in comparable ways. 

 % Within test relationships 
 %    (are automated scores related to scores on other sections of the test in similar ways compared to human scores?),
 % External relationships 
 %    (are automated scores related to other external measures of interest in similar ways compared to human scores?),
 % Relationship at the task type and reported score level
 %    (are the relationships similar at the task type and reported score level?)

\textit{Generalizability of scores} (4) questions the applicability of a single system to different tasks by analyzing facets (e.g., task types, prompts, genres, or populations of test-takers). It corresponds to the \textit{generalization} area of the framework.
It includes analyses with other scores than the gold standard, such as a single rater's score or an alternative aggregation of two human raters' scores. No specific metric is recommended in the original framework.
% , they discussed 
% the generalizability of scores across tasks and test forms, 
    % (How generalizable are the automated scores across tasks and test forms in comparison to human scores? 
    % how generalizable are the automated–human combined scores across test forms?), 
%    and
% the prediction of human scores on an alternate test form.
    % (To what extent do automated, human, and automated–human combined scores on one test form predict human scores on an alternate form?)

\textit{Score use and consequences} (5) is specific to the implementation of an AES system in a high-stakes environment (i.e., \textit{utilization}). The authors suggest analyzing the aggregated reported scores in depth to measure the impact on the final decision errors compared to human scores. Again, there is no specific recommended approach. However, the recommendation is to examine how automated scoring affects the accuracy of decisions, the claims and disclosures necessary for score users to ensure appropriate score interpretation, and the impact of automation on the test environment. In addition, the effects of automation should not be solely analyzed globally, but also across different subgroups to assess the fairness of the AES system.

% \textbf{references:} ``on ne peut pas dire que nous utilisons ces valeurs comme une norme.''
% \begin{itemize}
%     \item Standards for Educational and Psychological Testing  (2014):  \url{https://www.testingstandards.net/uploads/7/6/6/4/76643089/9780935302356.pdf}
%     \item Guidelines for Technology-Based Assessment (2022)  : \url{https://www.intestcom.org/upload/media-library/tba-guidelines-final-2-23-2023-v4-167785144642TgY.pdf}
% \end{itemize}

%\paragraph{Extrait : guidelines}
%Typically, the criteria for agreement between human and automated scoring are defined by relative orabsolute thresholds. A relative threshold would specify, for example, that the exact agreement of the
% engine with a human rater should be no more than 5\% lower than the exact agreement of two human
% raters with each other. Another example would be that 90\% of human and automated scores are within
% one score unit. An absolute threshold would specify, say, that the exact agreement of the engine with a
% human rater should be no lower than 70\%. A core psychometric principle is fairness. The performance of
% automated scoring systems should be evaluated for test taker subgroup populations to ensure scoring
% consistency holds across groups of test takers defined by gender, race/ethnicity, and other personal
% characteristics, including individuals with disabilities. (See also Chapter 10. Fairness and Accessibility).

\section{Evaluation Framework Methodology} \label{sec:Methodology}

% \st{In this work, we experimented with 8 AES architectures, presented in Section \ref{sec:models}, which were trained using the corpus described in Section \ref{sec:corpus}. For each architecture, we trained models using stratified 10-fold cross-validation\footnote{10\% of the dataset is used for validation. Therefore, we trained 10 models for each architecture with a dataset split in 80/10/10.}.} % For the evaluation using the homogeneous corpus, we considered just the test part of the cross-validation.

To assess our models, we extend the ABV framework proposed by \citet{williamson2012framework}, designing our evaluation around the five areas they outlined (introduced in Section \ref{sec:ConstructValidity}). Each area is addressed through one or more automated scores (Section \ref{sec:mapping}) that either follow or extend \citet{williamson2012framework}'s framework.
%by associating types of automated scores that might be used for assessing the five areas. 
%It is important to note that alternative approaches, such as the one by hand proposed by \citet{williamson2012framework}, remain valid. 
This section aims to pave the way for an automated assessment of AES systems based on the argument-based validity (ABV) framework, to improve the evaluation protocols of future AES systems. 
%Therefore, the gap between education, language, and NLP experts in this domain will be reduced.
% However, pointing out score types is insufficient for a reproductive evaluation framework.  Therefore, we propose the mapping between specific metrics and the five areas (Section \ref{sec:mapping}).
% With this methodology, we establish an automated AES assessment framework. A graphic mapping summary is shown in Table \ref{tab:summary}. 
The mapping between specific metrics proposed in this work and the five areas from the ABV framework is described in the rest of this section and is summarized in Table \ref{sec:mapping}.

% \begin{table*}
% \centering

% \begin{tabular}{p{6cm} | p{3cm} | p{6cm}} 
% \textbf{Emphasis Area}  & \textbf{Corresponding Inference in an Argument-Based Validity Framework} & \textbf{Automated Scores} \\ \hline
% Construct Relevance and Representation: Evaluating the ﬁt Between the Capability and the Assessment & Explanation & \textit{(a) linguistic features correlation with human rater} by rubric, tasks and prompt \\ \hline % c
% Empirical Performance—Association with Human Scores & Evaluation & \textit{(b) level order} (Spearman's correlation) and \textit{(c) label identification} (MSE, EA, AA and F1), and \textit{(d) agreement} (Cohen's Kappa and QWK) \\ \hline % a
% Empirical Performance—Association with Independent Measures & Extrapolation & \textit{(b) level order} and \textit{(c) label identification}, and \textit{(d) agreement} by raters' agreement \\ \hline % corpus quality + the big replacement
% Empirical Performance—Generalizability of Scores & Generalization & \textit{(b) level order} and \textit{(c) label identification}, and \textit{(d) agreement} on the Generalization Corpus \\ \hline
% Score Use and Consequences—Impact on Decisions and Consequences & Utilization & \textit{(f) fairness} (OSA, OSD, CSD) and \textit{(g) failure identification} (ECE and the relation between models' correctness and human agreement) \\ \hline % fairness + b + d

% \end{tabular}
% \caption{Alignment between framework and automated scores}
% \label{tab:summary}

% \end{table*}
\begin{table*}
\centering
\begin{tabular}{|p{6cm} | p{9cm}|}
\hline
\textbf{Emphasis Area}                                                                              & \textbf{Automated Scores}                                                                                                                                         \\ 
\hline
(1) Construct Relevance and Representation (explanation) %: Evaluating the ﬁt Between the Capability and the Assessment 
& \textit{(a) linguistic features correlation with human rater} by rubric, tasks and prompt                                                                         \\ 
\hline
% Empirical Performance—
(2) Association with Human Scores (evaluation)                                                & \textit{(b) ranking} (Spearman's correlation) and \textit{(c) label identification} (MSE, EA, AA and F1), and \textit{(d) agreement} (Cohen's Kappa and QWK)  \\ 
\hline
% Empirical Performance—
(3) Association with Independent Measures (extrapolation)                                        & 
out of scope for this study, as it requires additional scores 
% \textit{(b) level order} and \textit{(c) label identification}, and \textit{(d) agreement} by raters' agreement                                                   
\\ 
\hline
% Empirical Performance—
(4) Generalizability of Scores (generalization)                                                   & \textit{(b) ranking} and \textit{(c) label identification}, and \textit{(d) agreement} on the Generalization Corpus                                           \\ 
\hline
(5) Score Use and Consequences (utilization)
% —Impact on Decisions and Consequences  
& \textit{(e) fairness} (OSA, OSD, CSD) and \textit{(f) failure identification} (ECE and the relation between models' correctness and human agreement)              \\
\hline
\end{tabular}
\caption{Alignment between framework and automated scores}
\label{tab:summary}
\end{table*}

\subsection{Indicators for the ABV Framework} \label{sec:ConstructValidity}

%This section, along with the next, forms the core of our contribution. 
In this section, we report how we systematically expand on the ABV framework from \citet{williamson2012framework}, outlining our quantitative evaluation. 

%It is important to align our quantitative evaluation perspectives with the argumentative assessment practices in language \textcolor{orange}{testing}. We thus propose the following mapping with the framework from \citet{williamson2012framework}. 
%
%1) 

The \textit{(1) construct relevance and representation}, which measures the characteristics of the construct and the rubrics, naturally aligns with linguistic skills. While these skills can be precisely mapped to characteristics linked to the rubrics, this process becomes cumbersome in a holistic evaluation, as it requires annotating the corpus with these characteristics. Consequently, we opt for a more straightforward approach to approximate those, using automatically extracted linguistic features. %Thus, as these characteristics could be explained through the use of lexical and grammatical elements, linguistic features serve as a means to measure this aspect.
We study \ul{(a) the correlation of the linguistic features} in the following ways: 
correlation between linguistic features and CEFR levels in the scoring rubric assessment, 
correlation between linguistic features and CEFR levels across the tasks (in our case, three) in the construct assessment, 
and
correlation between linguistic features and CEFR levels for each prompt in the task design assessment. In language testing, the number of prompts is generally high and the number of essays per prompt low. Thus, to prevent statistical bias,we only retain prompts associated with more than 100 essays and compute the mean and standard deviation of the correlations. 
%Furthermore, in many cases, there are only a few examples of essays per prompt, which would cause a statistical bias. Therefore, this analysis only . 
% Finally, although we have presented evaluation measures that are linked to this area, it should be noted that this area also covers the assessment of the test's relevance as an evaluative tool (e.g. reporting goals), which is out of the scope of this paper.
%\textcolor{red}{Reporting goals - ??? etest design is our=t of scope
% (1.1) Construct evaluation - by task
% (1.2) Task design - prompts mais freq
% (1.3) Scoring rubric - general
% (1.4) Reporting goals - ???
%}

%2) 
The \textit{(2) association with typical scoring method} relates to model evaluation and is the most often reported area of the ABV framework \cite{williamson2012framework}.
In this work, we focus on three objectives: \underline{(b) ranking}, \underline{(c) label identification}, and \underline{(d) agreement}, all of which compare the system's predictions to the gold standard CEFR levels. The first one examines how the AES model ranks the essays,
% whether the AES model adequatly sorts the essays
while the second assesses how it assigns labels (e.g., CEFR levels) to essays. 
The last one evaluates the extent to which the model's predictions align with raters' judgments, correcting for agreement that might arise by chance.
%The last one addresses the consistency of the predictions with raters' judgments, accounting for the degree of agreement that could occur by chance. 
\citet{williamson2012framework} recommend using QWK. In addition, examples from the literature point out the need for other common metrics from classification and regression tasks (e.g., \citet{klebanovandmadnani2021}). These metrics align with recent AES practices with machine learning.
On top of this analysis on our entire %\textcolor{orange}{core}
corpus, we investigate how agreement between human raters impacts model performance.
To this end, we analyze subsets of our corpus (see Section \ref{sec:HomogeneousCorpus}). We propose three categories for sampling the subsets:
    \textit{full agreement} (human raters indicate the same score), % it targets a mix of large and reliable corpus for mainly train the AES models. The evaluation with this dataset assess the models' performance in the same setting they were trained on.
    \textit{close agreement} (difference of 1 level), and
    \textit{weak agreement} (difference of 2 levels).

%3) 
%Similar to construct relevance and representation, 
For \textit{(3) association with independent measures}, independent scores are necessary, as mentioned, to assess extrapolation. 
%external aspects of the essay, specifically its relationship with independent evaluation criteria (e.g., speech or reception proficiency scores in studies on written production proficiency, \textcolor{red}{or successful use of the language after the test}\textcolor{red}{VINCENT}). 
Since such additional scores are rarely available in AES research and published datasets and were not accessible for the TCF, this aspect falls outside the scope of this paper.
%relates to the essay's external elements, focusing on independent measures. As this extrapolation focuses on the comparison with additional scores usually not available in AES research, and they generally are not found in AES datasets either (e.g., speech or reception proficiency scores in written production proficiency AES studies), we do not include this area in the scope of this paper.
%uses tuning as a proxy for skill outcomes such as speaking (which we do not have access to in this article and which are not generally included in AES articles) 
%
% \subsection{Agreement and Generalization}
% As a plethora of different factors might impact the scoring process both at the human rater and the automatic systems, we investigate the impact of \underline{(d) agreement} between the raters in the model correctness as an indirect impact indication of AES performance. 
% For this agreement analysis, we scrutinize the corpus (see Section \ref{sec:HomogeneousCorpus}) to investigate the impact of human raters' agreement on automated systems. We propose three categories for essay sampling:
%     \textit{full agreement} (human raters indicate the same score), % it targets a mix of large and reliable corpus for mainly train the AES models. The evaluation with this dataset assess the models' performance in the same setting they were trained on.
%     \textit{close agreement} (difference of one level), and
%     \textit{weak agreement} (difference of two levels).

The \textit{(4) generalizability of scores} area assesses the applicability of the scoring system across different datasets. It evaluates performance under various test conditions (e.g., different tasks or different populations of test takers). 
%In this sense, we investigate AES performance according to the generalization of the results. 
For the generalization analysis, we assess the model's performance (i.e., \underline{(b) ranking abilities} and \underline{(c) label identification}, and \underline{(d) agreement}) on a corpus with unseen essays sampled to be representative of the target testing scenario, reflecting the model's application scenario. In addition, essays in this corpus have undergone a much more rigorous evaluation process, ensuring higher quality (see Section \ref{sec:HeterogeneousCorpus} for details).

%5) 
%The fifth area in the argument-based validity framework is utilization. 
The \textit{(5) score use and consequences} targets high-stakes environments and the potential for decision errors. 
To evaluate this area, we rely on two measures.
First, we study model \underline{(e) fairness} to identify biases regarding criteria unrelated to the scoring task (e.g., gender or native language). 
% This analysis assesses models' performance through the lens of the candidate's characteristics external to the task. 
We also assess \underline{(f) failure identification}, i.e., the possibility of identifying incorrect predictions of the model.

At this point, we have established six analysis categories corresponding to four out of the five areas of the ABV framework, with some categories applying to multiple areas. However, to produce an automatable assessment framework, these categories must be operationalized with concrete metrics, which we do in the next section.
%Therefore, in the following subsection, we define the concrete measures used within each category, ensuring that our evaluation remains both interpretable and reproducible.

\subsection{Construct Validity Metrics} \label{sec:mapping}

%\textbf{Construct validity}

Building on the six categories of metrics defined in Section \ref{sec:ConstructValidity}, this section introduces the concrete measures used within each category. This aims to ensure a reproducible evaluation protocol while preserving the interpretability of the original framework proposed by \citet{williamson2012framework}. 

%We propose four/six categories of analysis to assess construct validity, which are related to different construct's evaluation perspectives.
%
% \begin{itemize}
%     \item 
Most AES systems are automatically evaluated using direct comparison analysis, which assesses the relation between predictions and ground truth. % However, 
There is a tendency in the literature to report a single metric, QWK. As this may prevent capturing the full complexity of the models' performance, we follow \citet{kusuma2022automated}'s recommendation of using a wider range of evaluation metrics, aiming %. The goal is 
to minimize potential evaluation errors.
For the \underline{(b) ranking} metrics, we select the Rank Spearman's correlation ($\rho$), which evaluates only the order of predictions without considering the label. %, and Quadratically Weighted Kappa ($QWK$), which takes into account the rank order in addition to the standard agreement measures as Cohen's Kappa.
For the \underline{(c) label identification} metric, we follow \citet{klebanovandmadnani2021}, measuring the model's Exact Agreement ($EA$) for accuracy, Adjacent Agreement or Adjacent Accuracy ($AA$) for distinguishing between minor and severe errors, Mean Squared Error ($MSE$) for measuring the distance to the expected scores, and macro F\textsubscript{1} ($F1$) for assessing the balance between labels. %, and Cohen's Kappa ($\kappa$) evaluating the agreement between predicted and actual labels while accounting for chance agreement.
Although other metrics exist for this task, we chose these as they are most cited as best practices and studied in educational assessment or AES 
\cite{gwet2014handbook, lottridge2023psychometric, mccaffrey2022best, stemler2008best, von2014analyzing, yan2020validation}.
    % \item 

The \underline{(d) agreement} between the human and model labels relies on Cohen's Kappa ($\kappa$) and $QWK$ scores in this work. $\kappa$ evaluates the agreement between predicted and actual labels while accounting for chance agreement, while $QWK$ also considers the ordinal nature of predictions when computing the agreement. Note that $\kappa$ and $QWK$ could also be considered metrics of label identification and label ranking, respectively.

    % The second analysis, 
The \underline{(f) fairness} analysis aims to identify %possible 
    potential biases regarding candidates' characteristics that are unrelated to %that should not be taken into account for 
    the task (e.g., gender or prompt). The models' performance is examined through the lens of those characteristics. 
    Our fairness analysis follows the method proposed by \citet{loukina2019many}, evaluating three key metrics:
        overall score accuracy (OSA) assesses whether the model maintains equal accuracy for each group by comparing differences in squared error between human and automated scores;
        overall score difference (OSD) examines whether the model's predictions systematically differ from human scores for members of a certain group; and conditional score difference (CSD) verifies whether the model fairly assigns scores to candidates from different groups.
    %\textbf{metadata/sociolinguististic information??}.
    % \item

% \subsection{Agreement and Generalization}
% We propose to investigate AES performance according to both the generalisation of the corpus and the level of agreement between the raters. This results in a total of 5 analysis perspectives. We apply varying degrees of control when selecting essays for evaluation.  
% For the generalisation analysis, %we compare corpora with varying degrees of similarity to the training data.
% %Namely, 
% we assess the model's performance on a (1) Homogeneous Corpus, composed of essays sampled using the same criteria as the training data, 
% and on a (2) Heterogeneous Corpus, composed of additional essays sampled without any specific criteria.
% Those corpora are described in Section \ref{sec:corpus}.

% \subsubsection{Score Agreement}
% For the agreement analysis, 
% we scrutinize the Homogeneous Corpus for investigating the impact of human raters agreement on automated systems.
% We propose %the following 
% three categories for essay sampling:
% %
% %are divided into the following categories, from most restricted to less restricted:
% % \begin{itemize}
% %     \item 
%     (3) full agreement (raters indicate the same score), % it targets a mix of large and reliable corpus for mainly train the AES models. The evaluation with this dataset assess the models' performance in the same setting they were trained on.
%     (4) close agreement (difference of one level),
%     (5) weak agreement (difference of two levels).
    
    % Next, 
The \underline{(a) linguistic correlation} analysis aims to measure the degree of correlation (using Spearman's $\rho$) between %predicted 
    scores and linguistic features related to the construct. 
    These correlations help to assess how the models' predictions are aligned with the relevant underlying linguistic characteristics for language assessment. As correlation estimates can be computed at the rubric, task and prompt level, the number of data points may be insufficient to yield robust estimates (especially at the prompt level), as small sample sizes can introduce high variance. To mitigate this, we exclude cases with fewer than 100 essays and those where fewer than 5 of the 6 levels were represented. %This is important for correlations computed at the prompt level, where data sparsity can further impact reliability. 
    % \item 
    
Finally, the \underline{(f) failure identification} analysis aims to evaluate the model's trustworthiness.
    %\st{This means to what extent we can identify whether a model output is correct.} 
    %This analysis focuses on examining the model's suitability for unsupervised deployment scenarios. 
    To this end, we assess the models' calibration (i.e. the relation between the probability of a predicted level and its correctness) % of the level predicted by the model.    
            % For this assessment, we calculate
            by calculating the Expected Calibration Error (ECE) \cite{guo2017calibration}.
    Moreover, we analyze the relation between models' correctness and human agreement, dividing the corpus into cases where all models are correct, all models are wrong, or mixed. We report the average human agreement in each group. 

\section{Explored Models}
\label{sec:models}
For a comprehensive comparison of AES architectures, we follow the current preference for hybrid approaches in AES by combining linguistic features and deep learning. %As a baseline, we use two non-hybrid approaches, in order to assess the contribution of hybridization. As for the hybrid models, 
%\textcolor{blue}{In this work, we explored non-hybrid model following the work of \citet{wilkens-etal-2023-tcfle} on TCFLE-8 and hybrid models following the tendency in AES.} %
% As a baseline, we apply transformer-based models (TR) \textcolor{blue}{following the approach explored in the largest available corpus for French \cite{wilkens-etal-2022-cental}. %as they currently achieve state-of-the-art performance \cite{kusuma2022automated}. 
% Besides, we explore hybrid models through two ways of combining distributed representations and engineered features: explicit combination and implicit combination, based on multi-task learning.
 The models explored in this work were %already 
previously validated in the context of French readability assessment (a reception task) \cite{wilkens2024exploring}. Here, we extend their application to AES (a production task). While the tasks differ, the architectures align with state-of-the-art AES architectures.\footnote{Models' hyperparameters are discussed in Appendix \ref{app:hPrams}.}
%We detail those approaches below and %We considered the following flavors of implementation (
%outline them in Figure \ref{fig:archs}.  %\footnote{A comparison of these models in the context of readability assessment if presented in \textit{anonymous}.}
%In this work, we explored 8 different architectures\footnote{The range of hyperparameters and the selected values for each architecture are described in Appendix \ref{sec:appendix:Hyperparameters}.} (see Figure \ref{fig:archs}), which may be organized into three groups, based on the features integration method. A key element in the performance of these architectures is the linguistic features to be used. However, considering the different types of corpora explored in this work, it is natural to have different feature sets depending on the language and task. Therefore, the features are considered as a parameter for the architecture. %, and we apply a variable selection method in order to identify the most suitable set for each language and task pair.

% \begin{figure}
%     \includegraphics[width=0.48\textwidth, keepaspectratio]{hybrid-architectures}
%     \caption{Caption}
%     \label{fig:archs}
% \end{figure}

\begin{figure}[h]
  \begin{subfigure}{0.48\textwidth}
    \includegraphics[width=\textwidth, keepaspectratio]{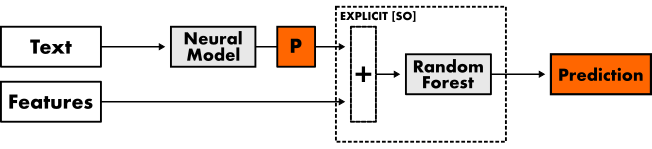}
    \caption{%Direct (or explicit) integration: 
    Soft-labeling} 
    \label{fig:arch_explicitSL}
  \end{subfigure}%

  \begin{subfigure}{0.48\textwidth}
    \includegraphics[width=\textwidth, keepaspectratio]{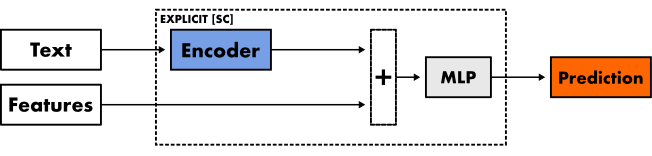}
    \caption{
    %Direct (or explicit) integration:
    Simple Concatenation} 
    \label{fig:arch_explicitSO}
  \end{subfigure}%

  %\hspace*{\fill}   % maximize separation between the subfigures
  
  \begin{subfigure}{0.48\textwidth}
    \includegraphics[width=\textwidth, keepaspectratio]{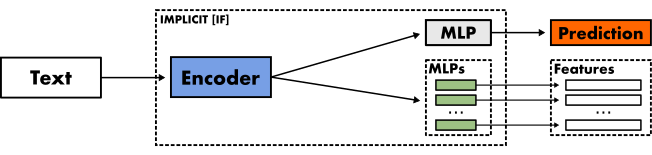}
    \caption{%Indirect (or implicit) integration: 
    Implicit Feature} 
    \label{fig:arch_implicitIF}
  \end{subfigure}%
  
\caption{The hybrid architectures} 
\label{fig:archs}
\end{figure}

\textbf{Baseline:} we use a transformer encoder (\textit{TR}) based on RoBERTa, more specifically CamemBERT \cite{martin-etal-2020-camembert}, as a non-hybrid baseline for measuring the contribution of the features. This model was previously explored for French AES by \citet{wilkens-etal-2023-tcfle} and \citet{sanchez-etal-2024-jingle} using the TCFLE-8 corpus. 

\textbf{Direct (or explicit) combination:} We explore two approaches. The first one leverages ensemble methods by using \textit{soft-labeling (SL)} and \textit{hard-labeling (HL)}. In both methods, the TR model is first fine-tuned, then its predictions are combined with features and used to train a Random Forest classifier (see Figure \ref{fig:arch_explicitSL}). With SL, the softmax output is used in the concatenation (see Figure \ref{fig:arch_explicitSL}), whereas only the final prediction %(i.e., \textit{argmax}) 
is used with HL. 
The second strategy which is similar to the one by \citet{uto2020neural}, the most common hybrid approach, % in AES, 
consists of feeding a multilayer perceptron (MLP) with the concatenation of %the document encoded by 
the TR's output (i.e. the CLS) and the features. % We considered the following flavors of implementation (exemplified in Figure \ref{fig:arch_explicit}).
We name this approach \textit{Simple Concatenation (SC)} (see Figure \ref{fig:arch_explicitSO}). %, simply combines the feature vector with the CLS vector, then this concatenated vector is fed to the output layer (i.e., MLP). In this architecture, the MLP is expected to be able to learn the target along with the mapping between the feature and transformer spaces.
Additionally, we can improve the %processing 
network capabilities by adding MLPs at different levels of the architecture. An MLP between the features and the concatenation is here named \textit{Concatenate MLP (CM)}, and a CM with an additional MLP after %between 
the encoder %and the concatenation 
is here named \textit{Concatenate 2xMLP (C2)}.
% \cutout{By adding an MLP between the features and the concatenation, we could simplify the task by allowing the network to separate the mapping between the spaces and/or even create a richer representation of the features. 
% This architecture, here named \textit{Concatenate MLP (CM)}, allows for greater exploration of the search space by adding a few more parameters to the network. % ($5 \times n$). In the scope of our work, we used an MLP with a first dense layer of $4 \times n$ neurons, followed by a dropout layer (10\%), followed by a dense layer of $2 \times n$ neurons that feeds a layer of $n$ neurons (output), where $n$ is the number of features.
% Following the same idea of including MLPs, the latest variant of the concatenation architecture, \textit{Concatenate 2xMLP (C2)}, also adds an MLP between the encoder output and the concatenation. Thus, the concatenation is performed on the output of two MLPs.}

\textbf{Indirect (or implicit) integration:} Language features can also be imprinted on the network through the use of auxiliary tasks, following a multi-task approach. %Here, we tested this idea by exploiting the same features used by the concatenation and RF architectures. Alternatively, we could exploit classic NLP tasks as proposed by \citet{zhou2019limit}, but this would prevent us from controlling indirect features learned by other tasks, making the comparison between the architectures unfair, as this architecture would have access to different information.
The first implicit architecture we explore, referred to as Implicit Features \textit{(IF)}, learns each linguistic feature as an independent regression task using an MLP (see Figure \ref{fig:arch_implicitIF}).\footnote{we apply a weight of 0.5 for the loss associated with the target task and 0.5 for the sum of all the feature regression losses.} %Thus, the network has $n+c$ output layers (where $c$ is the number of output neurons of the target task; in a regression $c = 1$). Since $n$ can vary depending on the corpus and can have a value considerably higher than $c$, the network could easily overlook the target task. 
% \cutout{In order to avoid a dominance of the linguistic features, we considered a weight of 0.5 for the loss associated with the target task and 0.5 for the sum of all the other losses.} %\textit{IF} assumes independence between features, which is not always required. We therefore proposed a simple variation of this architecture to exploit this aspect. 
A variant of IF, named \textit{Implicit Feature Vector (IV)}, groups all features into a single output vector. 
%See Figure \ref{fig:arch_implicitIF} for \textit{IF} and \textit{IV} architectures.

\subsection{Linguistic Features}
\label{methodo:feat}
% In view of the lack of studies combining linguistic features and language models for French AES, 
To compute the linguistic features needed for our analyses, we used the FABRA toolkit \cite{wilkens-etal-2022-fabra}, originally designed for readability assessment but also including features relevant to AES. FABRA identifies over 400 linguistic phenomena that are parameterized into features using 18 statistical aggregators, resulting in a total of 5,986 features\footnote{The entire list of variables is available at \url{cental.uclouvain.be/
fabra/docs_expert.html}}).
%The feature selection was made on the full sample of 27416 texts, which was split 10 times into train-dev-test for model experiments. We kept the 10 folds and used only the train part of each fold for this exploration of the features.
%The features were selected from FABRA, a toolkit based on the aggregation of a large number of variables predictive of a text’s readability in French. There are 18 different aggregators and over 400 variables, giving 5986 features as combination of variable/aggregator. 
%In this toolkit, language variables are grouped into categories (i.e. by type of linguistic phenomena such as lexical diversity groups or types of relation between types and tokens). 
%A features importance analysis was necessary in order to identify a set of features as diversified as possible while retaining significant predictive capacity. For diversity, we relied on the FABRA features classification into 6 groups (length based, lexical variables, syntactic variables…), each one containing families (e.g for lexical variables : content overlap, lexical frequency, affixes...) for a total of 27 families. We carried out the analysis family by family in order to select variables reflecting all aspects of the text. 
%
After annotation, 
%In order to understand the predictive power of the features,
we computed Spearman's, Pearson's and biserial correlations as well as Mutual Information Gain measures for each %FABRA 
feature, using the essay's level as the target. 
%We also computed point biserial correlation (comparing one level against all the other ones) in order to detect the particularities of each level as well as between each adjacent levels, in order to feed the model with interesting features for differentiating between two levels.
%These computations were performed on each "train" part of the 10 folds and we looked at two measures : the mean of the 10 results and the number of times a feature came as the most significant one in its FABRA family (max 10). 
%
% \begin{figure}[h!]
% \centering
% \includegraphics[width=0.5\textwidth]{latex/exemple_features.png}
% \caption{Example of results for the length based features across all levels.}
% \end{figure}
%
% Next, this
This information was analyzed by an expert FFL teacher, highly familiar with corpus annotation and feature engineering. The best features were then selected according to their relationship with essay level, pedagogical relevance, and informativeness. %contrast between features. 
% For each feature, the expert selection followed four criteria: %looked at the features candidates for each feature category, 
%  %and manually selected features based on the following four criteria: 
%  (1) the best correlation across all levels; (2) the best correlation for one level against the other or between adjacent levels; (3) the absence of overlapping information with another feature -- for example, learners essays can contain lexical and syntactic errors that are identified by specific variables targeting errors, but also affects the computation of features based on lexical frequency; (4) the robustness of the information represented by the aggregator. %For example, we favored the 10th and 90th percentiles, and 1st and 3rd quartile over min and max aggregators. 

\section{Corpus}
\label{sec:corpus}
%\st{The AES literature provides information for choosing a model, but AES for certification exams is a task that has been rarely explored in the literature. Furthermore,} 
% As discussed above, 
Corpora for language testing research are scarce, especially for French \cite{wilkens-etal-2023-tcfle}. %, as highlighted in Section \ref{sota:limits}.
In this work, we leverage access to an extensive collection of professionally-rated texts %written by learners and evaluated by professional raters in order 
to investigate %the development of 
AES for French as a foreign language (FFL) certification. %We collected essays
The data we use are collections of essays coming from one of the main French certification tests, the French knowledge test (TCF), run by \textit{France Education International}\footnote{\url{www.france-education-international.fr}} (FEI). \textcolor{update}{In 2024, candidates who sat for the computer-based TCF came from 88 countries and 350 test centers, representing 207 languages. A major challenge in automating a certification test with such global reach is test-taker heterogeneity. Some learned French as a foreign language, others as the language of schooling, and still others as a second or even a first language. 
% The status of French among candidates from the Francophonie is unclear and often lies between two learning types (e.g., school language and second language). 
%\textbf{They have very different native languages and educational backgrounds, and collecting data on these is impossible due to sensitivities in several countries' authorities.} 
TCF is available in several versions: a version targets  
the general public, while others target different aims such as naturalization and residence, and
immigration to Canada. To some extent, the versions embody the diversity of profiles. 
% For candidate evaluation, the French variety used is mainly a neutral French, often based on France’s variety but allowing for others. Candidates using vocabulary from the Francophonie are not penalized, provided the words are widely understood by Francophone speakers.
As regards the evaluation of candidates’ proficiency within the TCF, written performances are always independently double-rated, with a third rater if there is a discrepancy of more than one CEFR level. Research shows that for an exam with three tasks, double rating ensures sufficient reliability \cite{bouwer2015effect}. %; ideally, triple rating would be used \cite{bouwer2015effect}, but it would significantly increase candidate costs.
% To balance cost and reliability, FEI implements triple rating only when necessary—i.e., when there is insufficient agreement between markers.
}

% Reference: Bouwer, R., Béguin, A., Sanders, T., & Van Den Bergh, H. (2015). Effect of genre on the generalizability of writing scores. Language Testing, 32(1), 83‑100. https://doi.org/10.1177/0265532214542994

From TCF essays, two datasets were compiled for two different purposes: the core corpus and the generalization corpus. This section introduces both of them. In both datasets, there was no control for the task type and prompt that were used, which means that a variety of types and prompts are represented (based on random selection). This is to ensure that the results are not skewed towards a specific setting with regard to those aspects.

\subsection{Core Corpus} \label{sec:HomogeneousCorpus}

To build up our core corpus, we started from a raw dataset of 450,000 TCF essays provided by FEI~\footnote{Their availability for research has been granted through an agreement with FEI, including aspects relative to GDPR compliance. Unfortunately, the terms of use prevent us from publicly releasing the corpora used in this study.}. This dataset underwent multiple levels of cleaning. We followed the approach of \citet{wilkens-etal-2023-tcfle} to decide which outliers to remove and how % \cutout{First, we consider as outlier invalid essays (e.g. below A1 level, duplicate of the prompt, too short/long, or off-topic) and those whose scores adjusted by the Rasch method indicated a large difference from the original level (a standardized residual of 4 or more). %\footnote{In our empirical evaluation, we explored four standardized residue values (2, 3 and 4), observing that around 5.6\% of the corpus has a standardized residue of 2, 0.9\% has a value of 3 and 0.4\% a value of 4.}
% Next, we removed all cases where both raters disagree with each other plus with the candidate's final score. We also removed the candidates for which a distance of at least three CEFR levels was observed between the lowest and the highest ratings over the three tasks.
% }
to set the essay gold scores. \textcolor{update}{This score is determined based on a combination of the candidate's overall performance across the three written tasks of the TCF (narrative, descriptive, and argumentative) -- i.e. the candidate's CEFR level -- assigned by the raters. More precisely, for a given essay, the assigned score is the candidate's CEFR level when at least one of the raters also assigned that level to the essay. Alternatively, if both raters agreed on the essay's score, we duly assigned this level to the essay. Any essay that did not meet either of these conditions has been removed. It is important to emphasize that this French corpus is labeled according to a candidate-based evaluation, contrary to most available AES corpora. Therefore, the level is assigned to the candidate by combining the scores of each essay. The essays do not have a single CEFR score; instead, they have two (in some cases three), assigned by different raters.}

%This score is determined based on a combination of the candidate's overall performance across the three written tasks of the TCF (narrative, descriptive, and argumentative) -- i.e. the candidate's CEFR level -- and the raters' scores for the target task. More precisely, for a given essay, the score assigned is the candidate's CEFR level when at least one of the raters also assigned that level to the essay. Alternatively, if both raters agreed with the essay's score, we duly assigned this level to the essay. Any essay that did not meet either of these conditions has been removed.
%
%%%%%%%%%%%%%%%%%%%%%%%%%%%%%%
%%%%%%%%  sampling 
%%%%%%%%%%%%%%%%%%%%%%%%%%%%%%
%After the outlier removal step,
Next, we extracted a stratified random sample from the corpus, balancing the distributions of CEFR scores, adjusted scores\footnote{%Measuring writing skill implies considering various facets: candidate proficiency, rater leniency/harshness and difficulty of the task. To this aim, 
An adjusted version of the essay scores were produced through ``Many-facet Rasch measurement'', a psychometric approach that establishes a coherent framework for drawing reliable, valid, and fair inferences from rater-mediated assessments, thus answering the problem of fallible human ratings \citep{eckes_t_quantitative_2009}.}, and the language of use (the language most frequently used by the learner in daily life).  % of habitual use of the candidate\footnote{The language of habitual use is the language the candidate indicates as the one they usually use in everyday life.}. 
For sampling purposes, we considered only the most frequent candidate languages in the dataset (English, Arabic, Spanish, French\footnote{%A considerable amount of 
Candidates may indicate that their language of habitual use is French. This might include individuals living in French-speaking countries or native speakers who need to attest their proficiency in French.
}, Kabyle, Portuguese, and Russian), while the others were grouped into a category named \textit{Other}. In contrast to TCFLE-8, restricted to 8 languages of use, our dataset contains over 164 languages, providing a more comprehensive and representative depiction of the authentic situation of AES for French high-stakes testing.
This resulted in a dataset of 27,683 essays -- henceforth the \textbf{core corpus} -- with a rather balanced distribution over the 6 CEFR levels and 852 distinct prompts\footnote{A complete description of the size of the core corpus can be seen in Table \ref{tab:app:corpussize} in the appendix.}. %, as shown in Table~\ref{tab:description_sampledesc:docs_cefr_lang}.
In this corpus, 14,141 essays were given the same level by both raters (hereafter \textit{full agreement}), 12,750 had a difference of one level (\textit{close}), and 791 had a difference of two levels (\textit{weak}).

\subsection{Generalization Corpus} \label{sec:HeterogeneousCorpus}
%\paragraph{CORPUS VINCENT}
%The Test de connaissance du français (TCF) is a high-stakes test involving thousands of candidates. As such, its validity and reliability to support inferences based on the score must be carefully scrutinised. To meet this challenge applied to automated rating, 
FEI built an additional corpus of 961 finely calibrated essays -- henceforth the \textbf{generalization corpus}. It has been shown that using a carefully crafted corpus helps as a source of validity evidence to the evaluation of AES systems \cite{loukina-etal-2018-using}.
This generalization
corpus comprises essays sampled to represent as much as possible the distribution of the real-world scoring scenario of the TCF.
%\footnote{The \textcolor{orange}{generalization} corpus only considers that outliers have been removed.}.
%This corpus, contrary to the Homogeneous Corpus, is composed of essays related to those 
%The purpose of this corpus is to address the quality of the measurement, the generalisability of the results and the robustness of the various models developed to automate the assessment (Wind, 2018, Yan \& Bridgemen, 2020). 
%\cutout{To build it, an initial pool of 6 raters selected 1,083 writing performances. The selection was based on the content of the performances and their relation to the levels assessed by the raters during the operation. During the selection,}
%\st{In this corpus, 
%the raters registered their level of confidence between the content of the performance and the rated level to improve the validity of the selected content.} %After this initial step, 
To maximize the reliability and accuracy of the gold scores of this corpus, each essay was rated by a large number of raters: between 9 and 55 raters (depending on inter-rater agreement). Such a process was needed because of deviations in human ratings \cite{klebanovandmadnani2021}. 

\section{Results}
\label{sec:expe}

We experimented with 8 AES architectures (Section~\ref{sec:models}), all trained using the core corpus (Section~\ref{sec:corpus}). For each architecture, we trained models using stratified 10-fold cross-validation\footnote{10\% of the dataset is used for validation and 10\% for test. Consequently, 10 models were trained for each architecture with an 80/10/10 dataset split.}. 
In this section, we evaluate these 8 models based on the revised framework and all metrics described in Section~\ref{sec:mapping}. We thus present the correlation results (Section~\ref{sec:annotation}), the direct comparison analysis including ranking, label identification and agreement (Section~\ref{res:comp}), the fairness study (Section~\ref{sec:Fairness}), and the error identification study (Section~\ref{sec:FailureIdentification}).  % For the evaluation using the homogeneous corpus, we considered just the test part of the cross-validation.

\subsection{Linguistic Correlation Analysis} \label{sec:annotation}

We start by reporting interesting findings regarding the behavior of the 48 features selected as a result of the process described in Section \ref{methodo:feat}.\footnote{The list of features, with their correlation, is shown in Table~\ref{tab:selected_features}, Appendix \ref{app:feat_selection}.}
Among this selection, ``occurrence of errors" shows the highest correlation with the gold learner proficiency levels ($\rho = -0.79$), followed by lexical features capturing word diversity (0.64), the proportion of A1 words in the text (-0.61) and the average orthographic Levenshtein Distance to the 20 closest neighbors 
\cite{yarkoni2008moving} (0.57). 
The traditional word length feature -- computed as the number of syllables per word -- reaches 0.60, while sentence length has a lower correlation (0.43). Among the syntactic-based features, sentence depth is the most correlated one (0.52). 
The relatively strong correlations confirm that these linguistic features are well-suited to capture written proficiency, aligning with the goals of the assessment. Although the remaining selected features have a correlation coefficient below 0.5, they were retained in our models in order to diversify the information.

% \textcolor{blue}{\st{To the best of our knowledge, the features discussed in this section constitute the largest corpus study of a large set of linguistic features for French as a foreign language. In this paper, we used them to provide information for machine learning models. However, these results could also inform studies into the teaching and assessment of French as a foreign language.}}

Next, \textcolor{update}{to further study the associations between linguistic features and gold proficiency levels, we examine how these correlations evolve when the proficiency levels are those of a single rater (human or model}). For each rater and each feature, 
we compute the $\Delta$ between the correlation on the gold and the correlation of the given rater.
%we examine the association between the linguistic features and the models' predictions (presented in Section \ref{res:comp}). We focus on the strength of these associations by examining the difference between the correlation of linguistic features and the gold standard, on the one hand, and with raters (human and models), on the other hand.
Table~\ref{tab:results} (column $\Delta$ \textbf{Corr}) shows the results of this analysis.

On the core corpus, we observe that all raters have a minimal delta between their correlations. The human raters show an average difference of 0.01 for all features, while the models have values of about -0.01 (standard deviation $<$ 0.001 in all cases). This indicates a high level of stability in relation to features among all raters. The models tend to be slightly more aligned with the features than the human raters.
We repeat this procedure at both the task and prompt level. We observe a similar $\Delta$ between correlations across the 3 tasks. Regarding prompts, we exclude all cases with fewer than 100 essays (see Section \ref{sec:ConstructValidity}) and focus on the remaining 54 prompts, where we observe the same general trend as in the larger corpus. Human raters as well as the transformer (TR), the \textit{simple concatenation} (SC) and the \textit{concatenate MLP} (CM) models produce similar results, while the \textit{concatenate MLPx2} (C2), and both implicit models (IF and IV) show a slight reduction of the $\Delta$. The \textit{hard-labeling} architecture (HL) exhibits the largest difference (-0.03). These results indicate that task and prompt effects are independent in the models. 

Focusing on samples grouped by raters' agreement (\textit{full}, \textit{close}, and \textit{weak}), the average differences between humans and models remain comparable. Additionally, there is a clear increase in the mean differences as agreement decreases.

% \subsection{Fairness Evaluation Analysis}

% In the fairness analysis, we found no bias towards usual language, gender, or test task with any of the three metrics. When grouping the samples of the \textcolor{blue}{large} corpus by the degree of agreement between the annotators, we did not identify any gender bias. Regarding task and language, we observed a small bias %(about 0.13 of $R^2$) 
% in the weak agreement sample. %\footnote{
% The usual language bias identified is 
% 0.20 of $R^2$ in OSA for \textit{C2},
% 0.15 in OSD for \textit{CM}, and
% 0.11 in CSD for \textit{C2}.
% For task bias, the OSA metric identified 0.1 of $R^2$ in \textit{HL}, and 0.13 (\textit{IF}), 0.11 (\textit{IV}) and  0.12 (\textit{SL}) in the OSD.
% %} 
% However, it is important to highlight that this bias was identified in models from only one repetition out of ten. Finally, we assessed fairness in the \textcolor{blue}{generalization} corpus, where no bias was identified.
% %In addition, we explored the evaluators' agreement by language, where no bias between languages was observed.

\subsection{Models Performance} % Classification Metrics / Direct Comparison Analysis
\label{res:comp}

% %\subsection{Study 1}
% After selecting the features, we can move on to comparing the different architectures explored in this work. 
% For this evaluation, we used a 10-fold cross-validation, and % using the same stratification discussed in Section \ref{sec:corpus}, resulting in splits of 80/10/10 for training, development and testing. The
% the result of this evaluation is displayed at column ``full'' in Table \ref{tab:resultsF1}.

% \textcolor{blue}{In this section, we report and analyze our models' performance following three steps: (1) standard evaluation on the core corpus; (2) evaluation on the generalization corpus; and (3) quantitative analyses of the best model.}

\subsubsection{Standard
evaluation on the core corpus}
% First, 
The results of the Direct Comparison Analysis are shown in Table~\ref{tab:results}, where we can observe a remarkable stability. Only small performance differences are associated to the architectures, which confirms the findings of \citet{rodriguez2019language} and \citet{mayfield2020should} about BERT efficiency. 
When considering the entire core corpus, the automated models fall behind the human raters\footnote{
We compute the agreement between the raters, using Cohen's kappa and QWK. In this corpus, a large number of raters participated in the annotation, but for the sake of simplicity, we do not distinguish individuals and compare ``rater 1'' and ``rater 2'' for each text, obtaining an agreement of 0.4  for $\kappa$ and 0.83 for QWK. This agreement is similar to those observed in the literature \cite{taghipour2016neural}. 
%
% The raters can predict the final CEFR score of an essay with an acceptable F1 of 0.65 and 0.71 of QWK. However, the errors of the evaluators are rarely above one level as their adjacent accuracy is 0.99. We do not observe an impact of the native language of candidates on the raters' performance.
} as regards $\kappa$, $QWK$, $EA$, and $F1$. 
A closer examination of the models' performance and raters' agreement reveals a strong performance of human raters. 
This result is partly explained by the fact that their scores contributed to the creation of the gold standard. Interestingly, this bias disappears when comparing cases of full agreement with those of weak agreement, where human raters actually perform worse than the models in $\rho$, $QWK$, $MSE$ and $AA$ (see Table~\ref{tab:results}). 
Moreover, comparing the metrics that penalize all errors equally (e.g., $F1$ and $\kappa$) and those that penalize errors involving adjacent levels less (e.g., $QWK$, $AA$, $\rho$), we see that most model errors are off by one level (e.g., A1 or B1 instead of A2).

Focusing on the architecture of the models, we observe in Table~\ref{tab:results} that Transformer (\textit{TR}) is competitive with the hybrid models. Although the top-performing models vary depending on the combination of corpus sample and the evaluation metric, the \textit{soft-labeling} strategy (\textit{SL}) tends to perform the best on the core corpus as well as on samples with full and close agreement. \textit{Simple concatenation} (\textit{SC}) is slightly better on the weak agreement sample.

As regards the explicit hybrid models, they do not differ statistically from each other on the core corpus, even though \textit{SC} seems to perform slightly better. On the contrary, the implicit models differ statistically from each other (\textit{IV} achieves better performance), and 
the ensemble models (\textit{SL} and \textit{HL}) too. 
If we only consider the top model from each architecture family, the \textit{soft-labeling} strategy (\textit{SL}) outperforms the other architectures, followed by \textit{simple concatenation} (\textit{SC}) and Transformer (\textit{TR})\footnote{\textit{SC} and \textit{TR} do not statistically differ from each other.}. %The best models in each family are highlighted in bold in Table \ref{tab:results}.

\begin{table*}[t!h!]
\centering
\begin{tabular}{|rl|c|r|r|r|r|r|r|r|r|} 
\hline
\multicolumn{2}{|l|}{\multirow{2}{*}{\textbf{Corpus}}}                                        & \multirow{2}{*}{\textbf{Model}} & \multicolumn{7}{c|}{\textbf{Direct Comparison}}                                                                                                                              & \multicolumn{1}{c|}{\multirow{2}{*}{\begin{tabular}[c]{@{}c@{}}\textbf{$\Delta$}\\\textbf{Corr}\end{tabular}}}  \\
\multicolumn{2}{|l|}{}                                                                        &                                 & \multicolumn{1}{r}{$\rho$} & \multicolumn{1}{r}{$\kappa$} & \multicolumn{1}{r}{$QWK$} & \multicolumn{1}{r}{$MSE$} & \multicolumn{1}{r}{$EA$} & \multicolumn{1}{r}{$AA$} & $F1$               & \multicolumn{1}{c|}{}                                                                                    \\ 
\hline
\multicolumn{2}{|l|}{\multirow{9}{*}{\begin{tabular}[c]{@{}l@{}}Core\\Corpus\end{tabular}}}   & TR                              & 0.90                    & 0.48                      & 0.89                    & 0.50                    & 0.57                   & 0.98                   & 0.57             & -0.02                                                                                                    \\ 
\cline{3-11}
\multicolumn{2}{|l|}{}                                                                        & CM                              & 0.90                    & 0.48                      & 0.89                    & 0.49                    & 0.57                   & 0.98                   & 0.57             & -0.02                                                                                                    \\ 
\cdashline{3-11}
\multicolumn{2}{|l|}{}                                                                        & C2                              & 0.90                    & 0.48                      & 0.89                    & 0.49                    & 0.57                   & 0.98                   & 0.57             & -0.02                                                                                                    \\ 
\cdashline{3-11}
\multicolumn{2}{|l|}{}                                                                        & SC                              & 0.90                    & 0.44                      & 0.88                    & 0.55                    & 0.54                   & 0.97                   & 0.54             & -0.02                                                                                                    \\ 
\cline{3-11}
\multicolumn{2}{|l|}{}                                                                        & IF                              & 0.90                    & 0.48                      & 0.89                    & 0.48                    & 0.58                   & 0.98                   & 0.57             & -0.02                                                                                                    \\ 
\cdashline{3-11}
\multicolumn{2}{|l|}{}                                                                        & IV                              & 0.90                    & 0.48                      & 0.89                    & 0.49                    & 0.57                   & 0.98                   & 0.57             & -0.01                                                                                                    \\ 
\cline{3-11}
\multicolumn{2}{|l|}{}                                                                        & SL                              & 0.90                    & 0.50                      & 0.89                    & 0.44                    & 0.60                   & 0.99                   & 0.59             & -0.03                                                                                                    \\ 
\cdashline{3-11}
\multicolumn{2}{|l|}{}                                                                        & HL                              & 0.90                    & 0.47                      & 0.88                    & 0.48                    & 0.57                   & 0.98                   & 0.56             & -0.02                                                                                                    \\ 
\cline{3-11}
\multicolumn{2}{|l|}{}                                                                        & Raters                          & 0.91                    & 0.70                      & 0.91                    & 0.40                    & 0.75                   & 0.98                   & 0.75             & 0.01                                                                                                     \\ 
\hline \hline
\vcell{}         & \multirow{5}{*}{\begin{tabular}[c]{@{}l@{}}Full\\agreement\end{tabular}}           & \vcell{TR}                      & \vcell{0.92}            & \vcell{0.55}              & \vcell{0.91}            & \vcell{0.40}            & \vcell{0.63}           & \vcell{0.99}           & \vcell{0.64}     & \vcell{-0.01}                                                                                            \\[-\rowheight]
\printcellmiddle &                                                                            & \printcellmiddle                & \printcellmiddle        & \printcellmiddle          & \printcellmiddle        & \printcellmiddle        & \printcellmiddle       & \printcellmiddle       & \printcellmiddle & \printcellmiddle                                                                                         \\ 
\cline{3-11}
                 % &                                                                            & CM                              & 0.92                    & 0.56                      & 0.91                    & 0.39                    & 0.64                   & 0.99                   & 0.64             & -0.01                                                                                                    \\ 
% \cdashline{3-11}
                 % &                                                                            & C2                              & 0.92                    & 0.55                      & 0.91                    & 0.40                    & 0.63                   & 0.99                   & 0.63             & -0.01                                                                                                    \\ 
% \cdashline{3-11}
                 &                                                                            & SC                              & 0.92                    & 0.51                      & 0.90                    & 0.45                    & 0.60                   & 0.98                   & 0.60             & -0.01                                                                                                    \\ 
\cline{3-11}
                 % &                                                                            & IF                              & 0.92                    & 0.56                      & 0.91                    & 0.38                    & 0.64                   & 0.99                   & 0.64             & -0.01                                                                                                    \\ 
% \cdashline{3-11}
                 &                                                                            & IV                              & 0.92                    & 0.55                      & 0.91                    & 0.39                    & 0.63                   & 0.99                   & 0.64             & -0.01                                                                                                    \\ 
\cline{3-11}
                 &                                                                            & SL                              & 0.92                    & 0.57                      & 0.91                    & 0.36                    & 0.66                   & 0.99                   & 0.65             & -0.02                                                                                                    \\ 
\cdashline{3-11}
                 % &                                                                            & HL                              & 0.91                    & 0.56                      & 0.91                    & 0.38                    & 0.64                   & 0.99                   & 0.63             & -0.01                                                                                                    \\ 
\cline{3-11}
                 &                                                                            & Raters                          & 1.00                    & 1.00                      & 1.00                    & 0.00                    & 1.00                   & 1.00                   & 1.00             & -0.00                                                                                                    \\ 
\hline \hline
\vcell{}         & \multirow{5}{*}{\begin{tabular}[c]{@{}l@{}}Close\\agreement\end{tabular}}           & \vcell{TR}                      & \vcell{0.89}            & \vcell{0.41}              & \vcell{0.87}            & \vcell{0.59}            & \vcell{0.52}           & \vcell{0.97}           & \vcell{0.51}     & \vcell{-0.02}                                                                                            \\[-\rowheight]
\printcellmiddle &                                                                            & \printcellmiddle                & \printcellmiddle        & \printcellmiddle          & \printcellmiddle        & \printcellmiddle        & \printcellmiddle       & \printcellmiddle       & \printcellmiddle & \printcellmiddle                                                                                         \\ 
\cline{3-11}
                 % &                                                                            & CM                              & 0.89                    & 0.40                      & 0.87                    & 0.59                    & 0.51                   & 0.97                   & 0.50             & -0.02                                                                                                    \\ 
% \cdashline{3-11}
                 % &                                                                            & C2                              & 0.89                    & 0.42                      & 0.87                    & 0.56                    & 0.53                   & 0.97                   & 0.51             & -0.02                                                                                                    \\ 
% \cdashline{3-11}
                 &                                                                            & SC                              & 0.89                    & 0.37                      & 0.86                    & 0.65                    & 0.48                   & 0.96                   & 0.47             & -0.02                                                                                                    \\ 
\cline{3-11}
                 % &                                                                            & IF                              & 0.89                    & 0.41                      & 0.87                    & 0.57                    & 0.52                   & 0.97                   & 0.50             & -0.02                                                                                                    \\ 
% \cdashline{3-11}
                 &                                                                            & IV                              & 0.89                    & 0.41                      & 0.87                    & 0.57                    & 0.52                   & 0.97                   & 0.50             & -0.02                                                                                                    \\ 
\cline{3-11}
                 % &                                                                            & HL                              & 0.89                    & 0.45                      & 0.88                    & 0.50                    & 0.55                   & 0.98                   & 0.53             & -0.03                                                                                                    \\ 
% \cdashline{3-11}
                 &                                                                            & SL                              & 0.88                    & 0.40                      & 0.86                    & 0.56                    & 0.52                   & 0.98                   & 0.49             & -0.02                                                                                                    \\ 
\cline{3-11}
                 &                                                                            & Raters                          & 0.87                    & 0.40                      & 0.86                    & 0.61                    & 0.51                   & 0.99                   & 0.50             & 0.01                                                                                                     \\ 
\hline \hline
\vcell{}         & \multirow{5}{*}{\begin{tabular}[c]{@{}l@{}}Weak\\agreement\end{tabular}}           & \vcell{TR}                      & \vcell{0.84}            & \vcell{0.23}              & \vcell{0.82}            & \vcell{0.87}            & \vcell{0.36}           & \vcell{0.92}           & \vcell{0.36}     & \vcell{-0.05}                                                                                            \\[-\rowheight]
\printcellmiddle &                                                                            & \printcellmiddle                & \printcellmiddle        & \printcellmiddle          & \printcellmiddle        & \printcellmiddle        & \printcellmiddle       & \printcellmiddle       & \printcellmiddle & \printcellmiddle                                                                                         \\ 
\cline{3-11}
                 % &                                                                            & CM                              & 0.84                    & 0.21                      & 0.82                    & 0.86                    & 0.34                   & 0.94                   & 0.35             & -0.05                                                                                                    \\ 
% \cdashline{3-11}
                 % &                                                                            & C2                              & 0.83                    & 0.24                      & 0.82                    & 0.86                    & 0.37                   & 0.93                   & 0.36             & -0.05                                                                                                    \\ 
% \cdashline{3-11}
                 &                                                                            & SC                              & 0.83                    & 0.24                      & 0.82                    & 0.91                    & 0.37                   & 0.91                   & 0.37             & -0.06                                                                                                    \\ 
\cline{3-11}
                 % &                                                                            & IF                              & 0.83                    & 0.23                      & 0.81                    & 0.85                    & 0.36                   & 0.93                   & 0.34             & -0.05                                                                                                    \\ 
% \cdashline{3-11}
                 &                                                                            & IV                              & 0.83                    & 0.24                      & 0.82                    & 0.85                    & 0.37                   & 0.93                   & 0.35             & -0.05                                                                                                    \\ 
\cline{3-11}
                 &                                                                            & SL                              & 0.83                    & 0.21                      & 0.81                    & 0.84                    & 0.35                   & 0.94                   & 0.33             & -0.06                                                                                                    \\ 
\cdashline{3-11}
                 % &                                                                            & HL                              & 0.82                    & 0.18                      & 0.80                    & 0.88                    & 0.32                   & 0.94                   & 0.30             & -0.06                                                                                                    \\ 
\cline{3-11}
                 &                                                                            & Raters                          & 0.67                    & 0.33                      & 0.66                    & 1.88                    & 0.45                   & 0.57                   & 0.44             & 0.06                                                                                                     \\
\hline
\end{tabular}
    \caption{Results of the direct comparison and linguistic correlation analysis in the core corpus, including its splits by human raters agreement (the results for all the models are presented in Table \ref{tab:resultsFull}).}
    \label{tab:results}
\end{table*}

\textcolor{update}{A more fine-grained inspection of the results reveals that the models' scores tend to decrease as the CEFR level increases (see Table \ref{tab:f1by_level}), in contrast with human raters. %While the best score in a specific level is model-dependent, it is possible to identify some patterns.
While the best-performing model at a given level varies, some consistent patterns emerge. %\textit{Soft-labeling} (SL) achieves the best overall performance, statistically outperforming the others, as already seen before. 
All the models (except HL) perform best at the A1 and A2 levels, which contain texts with highly typical lexico-grammatical features (short sentences, simple words, etc.). From level B1 onwards, however, there is a general deterioration, which becomes more important and culminates at the C2 level for some models (C2, SC, IF, and especially SL and HL). 
% but deteriorates at the C2 level. The vanilla transformer (TR) remains rather constant after an initial drop between A2 and B1. 
%vanilla transformer (TR) model performs similarly to, or better than, the explicit concatenation models (CM, SC, C2) at the C levels, especially at C2.
A possible explanation for this performance drop at the C levels is that skilled written productions can be characterized by a wider set of linguistic properties (e.g., coherence and cohesion, implicitness, phraseology, etc.). It has been acknowledged in the learner corpus literature \cite{forsbergandlindqvist2014,declerqandhousen2017} that complexity features generally have a weaker discriminative power at C levels. 
Another analysis consists of comparing the performance of the transformer, which relies solely on the information contained in the embeddings, with that of the hybrid models that also integrate features. It is interesting to note that TR does not exhibit any performance deterioration at the C2 level, whereas the models that rely most heavily on the variables (in particular SL and HL) suffer the most. Conversely, these same models seem to benefit the most from the effect of the variables at the B2 and C1 levels. This effect is weaker or even absent for the explicit models, perhaps because the presence of MLP layers mitigates it.}

%\textcolor{blue}{as combined with efficient A-level features available to the concatenation models, while TR, without access to them, is better at generalizing the C levels.  Nevertheless, the C2 level is the hardest level for all models. 
% Moreover, HL and SL have a significant relative improvement in B2 and C1 when compared with the other models and a similar performance in the other levels (except for HL at A1 level), thus showing a difference when combining the features directly with the transformers (e.g., late fusion) compared with training independent models.
Looking at the classification errors of the best-performing model, the \textit{soft-labelling} strategy (SL), % (Figure \ref{fig:ConfMatrix}), 
we can see that it tends to assign a higher CEFR level when the expected level is B2 or below, while it assigns a lower level when the expected level is C1 or C2. 
Moreover, $\sim$56\% of the C2 essays are misclassified as C1, and $\sim$40\% are correctly classified as C2. This confirms the weakness of the model in identifying essays of level C2 compared to the gold. 

\begin{table*}[h!]
\centering
\begin{tabular}{|l|c|c|c|c|c|c|c|} 
\hline
\multicolumn{1}{|c|}{\textit{Model}} & \multicolumn{1}{c|}{$\rho$} & \multicolumn{1}{c|}{$\kappa$} & \multicolumn{1}{c|}{$QWK$} & \multicolumn{1}{c|}{$MSE$} & \multicolumn{1}{c|}{$EA$} & \multicolumn{1}{c|}{$AA$} & \multicolumn{1}{c|}{$F1$}  \\ 
\hline
TR                                   & 0.93                     & 0.56                       & 0.91                     & 0.42                     & 0.63                    & 0.99                    & 0.63                     \\ 
\hline
SC                                   & 0.93                     & 0.56                       & 0.91                     & 0.41                     & 0.64                    & 0.99                    & 0.63                     \\ 
\hline
IV                                   & 0.93                     & 0.56                       & 0.91                     & 0.43                     & 0.64                    & 0.99                    & 0.63                     \\ 
\hline
SL                                   & 0.93                     & 0.60                       & 0.92                     & 0.37                     & 0.67                    & 0.99                    & 0.66                     \\ 
\hline \hline % \hline
Rater1                               & 0.92                     & 0.51                       & 0.88                     & 0.58                     & 0.59                    & 0.96                    & 0.51                     \\ 
\hline
Rater2                               & 0.92                     & 0.58                       & 0.89                     & 0.51                     & 0.65                    & 0.96                    & 0.56                     \\ 
\hline \hline % \hline
TR\textsubscript{vote}                                  & 0.93                     & 0.56                       & 0.91                     & 0.41                     & 0.64                    & 0.99                    & 0.63                     \\ 
\hline
SC\textsubscript{vote}                                  & 0.94                     & 0.60                       & 0.92                     & 0.37                     & 0.67                    & 0.99                    & 0.66                     \\ 
\hline
IV\textsubscript{vote}                                  & 0.93                     & 0.57                       & 0.91                     & 0.42                     & 0.65                    & 0.99                    & 0.64                     \\ 
\hline
SL\textsubscript{vote}                                  & 0.93                     & 0.61                       & 0.92                     & 0.35                     & 0.68                    & 0.99                    & 0.67                     \\
\hline
\end{tabular}
\caption{Results of the direct comparison analysis in the generalization corpus.}
\label{tab:generalization}

\end{table*}

% Statsitical diff:
%          CS22HL  CS22SL  CS22cs0  CS22cs1  CS22cs2  CS22cs3  CS22cs4  CS22cs5
% CS22HL    False    True     True     True     True     True    False     True
% CS22SL     True   False     True    False    False     True     True     True
% CS22cs0    True    True    False    False    False    False     True    False
% CS22cs1    True   False    False    False    False    False     True    False
% CS22cs2    True   False    False    False    False    False     True    False
% CS22cs3    True    True    False    False    False    False     True    False
% CS22cs4   False    True     True     True     True     True    False     True
% CS22cs5    True    True    False    False    False    False     True    False

%\subsection{Study 2}

%\sout{Although our results do not show a clear positive impact when using hybrid models, studies using feature concatenation tend to report better performance over simple transformers. This difference may have different origins. Beside a possible publication bias, it could be related to the size or the language of the corpus, as well as to the difference between CamemBERT \cite{martin-etal-2020-camembert} and BERT models (which are normally used for English).}
\subsubsection{Evaluation on the generalization corpus}

Evaluating our models on the generalization corpus brings more insights. The results of the direct comparison metrics are presented in Table \ref{tab:generalization}. We only report on the best architecture from each family.
The most prominent result is the performance gain for the generalization corpus over the core corpus. Specifically, this difference tends to get smaller when only the full agreement sample is considered. Unlike the results on the core corpus, the human raters' performance decreases. This is explained by the already mentioned fact that the gold scores for the core corpus are based on the decisions of only two raters who are also used in our analysis, resulting in symmetric results when comparing each rater with the gold standard. On the other hand, the reference score of an essay is determined by a large pool of raters in the generalization corpus, thereby reducing the variation introduced by individual raters. Consequently, for comparison purposes, we reported raters' performance compared with the gold scores (Rater 1 and Rater 2) considering only two randomly selected raters out of the larger pool of raters for each essay. Comparing the performance of human and automated raters, we see that the models' performance is equal or superior to the humans' across all measures. This points out the excellent generalization capacity of the models and the critical need for such evaluation corpora to properly assess the performance of AES systems before they are deployed in production. Regarding model performance, \textit{SL} shows superior results across all metrics again. 

Regarding the models used to rate the essays of the generalization corpus, it is important to note that 10 different models trained with different samples from the core corpus were used (each model is trained on one of the 10 repetitions from the cross-validation procedure). However, it is also important to bear in mind that a common approach in AES consists in combining models trained on different samples (i.e., as proposed by \citet{taghipour2016neural}). Therefore, we also evaluate ensemble models using the majority vote for prediction.\footnote{Our aim is to assess an improvement in the combination of models, rather than to identify the best way of combining them. For a discussion of different methods of combining AES models, see \citet{uto2023integration}.} This means that, for each model, we predict the essay levels in the generalization corpus. In Table \ref{tab:generalization}, we indicate the results based on this approach with the \textit{vote} suffix. The voting approach mostly improved the models. However, the \textit{SL} architecture keeps its top position, even with the small improvement caused by the voting approach. This confirms the robustness of this architecture. Interestingly, the difference between \textit{TR} and \textit{TR\textsubscript{vote}} is negligible. 
\textcolor{update}{Despite the slightly superior performance of the voting model in this study, its use implies a significantly higher computational cost. Thus, our results point to the use of SL without voting.}

\subsubsection{Analyses of the best model}

We conducted a series of more in-depth analyses of the best-performing model, namely the \textit{soft-labeling} approach. As previously, we computed the model’s performance by CEFR level, but on the generalization corpus, which offers substantially more reliable assessments. Comparing the distribution of SL on the core corpus and the generalization corpus (line SL-GC in Table \ref{tab:f1by_level}), we see that results on A levels surprisingly drop, and dramatically increase for all other levels, including C2 (in contrast with previous findings on the core corpus). These results call into question our previous hypothesis regarding the limited discriminative power of our variables at the C level. It could be instead that human rating of C2-level essays may be especially difficult to perform consistently. A plausible contributing factor of this seems to be the underspecified nature of the CEFR descriptors at the C2 level\footnote{An example of this underspecification can be directly found in the Companion volume of the CEFR \cite[41]{council2020}: "For example, at C2 the entry is sometimes: No descriptor available: see
C1".}.

% Différence entre le test set et FEI validation :
%   0.55 (val)  0.62 (val) → globalement, le modèle se comporte également mieux sur des données dont le niveau est mieux estimé, via un consensus humain solide(lien avec les limites des descripteurs du CECR). 
%    Niveau C2 se comporte nettement mieux sur la validation → confirmerait qu’il y a bien un problème au niveau des évaluateurs humains dans le corpus de test.  
    
% les correcteurs n'ont pas tous une vision intégrée du niveau C2 (C2 construct is actually weakly-defined:   
% -> "Pas d'erreurs d'orthographe" dans le CECR au C2; language can be efficient but simple and is therefore not typical of a C2 level
% chercher une référence (c'est une des difficultés du français)

% Ajouter petit tableau avec comparaison core et generalisation

\textcolor{update}{
% As another exploratory experiment, 
We also studied the distribution of language errors in the essays and its relation to the prediction errors of the \textit{soft-labeling} model (i.e., residuals). For this, we manually analyzed errors made by the candidates in a sample of essays. We subsampled 308 essays in the test corpus, using stratified sampling based on two criteria: the CEFR levels of the texts and values of the residuals. %(i.e. the absolute value of the discrepancy between the reference CEFR and the predicted level). 
In each essay, we manually classified errors into 
%identified the errors and assigned them to 
three categories: lexical, (e.g., spelling mistakes, extra or missing diacritics, non-existing words), 
morphological (e.g., agreement or conjugation errors) 
and syntactic.  (e.g., extra or missing punctuation/words, word order issues). 
For each level, we then compute the Spearman's correlation %coefficient 
between the number of errors of each category 
(normalized over document length) and the model's residuals. 
% Table \ref{tab:corr_errors}, in Appendix \ref{app:corr_errors}, shows the number of documents of each level, and the resulting correlations along with their significance levels.
}

% REMI : lien entre la tokenization et les performances.

\textcolor{update}{Table \ref{tab:corr_errors:proportion} displays the proportion of each type of errors across the CEFR levels, indicating a clear decline in errors as proficiency increases. For the correlations between residuals and type of errors (see Table \ref{tab:corr_errors:corr}), the only significant ones can be found at low proficiency levels (A1 and A2) and C2 for the lexical category and at C2 for the syntactic category. Those suggest that the model may strongly rely on the presence of lexical errors to identify the A levels (as the model tends to assign a higher level to A-level texts with few errors), and on the absence of lexical errors to identify the C2 level (as C2 texts having more errors are more poorly classified). 
The text examples in Table \ref{app:tab:error} (Appendix \ref{app:model_errors}) illustrate this heavy reliance on errors: texts classified as B1 by humans based on fluency and syntax are penalized as A1 due to a large amount of lexical errors. We deem this phenomenon worthy of further investigation and leave it as future work, as it would require annotating a larger volume of data.}

\subsection{Fairness Evaluation Analysis}
\label{sec:Fairness}

The fairness analysis shows no bias towards language of use, gender, or task with any of the three metrics. When sampling the core corpus by the degree of agreement between the raters, no gender bias appears neither. Regarding task and language, a small bias %(about 0.13 of $R^2$) 
can be observed on the weak agreement sample. %\footnote{
The bias on language of use amounts to 
0.20 of $R^2$ for overall score
accuracy (OSA) for the \textit{C2} model,
0.15 for the overall score difference (OSD) for \textit{CM}, and
0.11 for the conditional score difference (CSD) for \textit{C2}.
For task bias in specific models, small biases were identified by the OSA metric (0.1 of $R^2$ in \textit{HL}) or the OSD metric (0.13 (\textit{IF}), 0.11 (\textit{IV}) and 0.12 (\textit{SL})).
%} 
However, %it is important to highlight that 
this bias is only identified in one out of ten models trained using the cross-validation. 
In addition, as a proxy for test-taker background, we analyze the performance of models on each of the TCF variants (e.g., Canada and French citizenship) and found no substantial differences. 
Finally and more importantly, no bias has been identified on the generalization corpus.
%In addition, we explored the evaluators' agreement by language, where no bias between languages was observed.

\textcolor{update}{The above quantitative fairness analysis was based on classic factors (e.g., gender, language and task). We also investigated complementary background-related factors, which can also shape models' outcomes. We looked at essays with severe classification errors (examples available at Table~\ref{app:tab:error}, Appendix \ref{app:model_errors}).
}
%
%
%
%We further investigated the fairness of our models through a qualitative textual analysis of classification errors. 
%We looked at texts where there was a big difference between human evaluators and the model (see Table \ref{app:tab:error} in Section \ref{app:model_errors} in the Appendix for the examples).
%We note that a set of essays was incorrectly classified, likely due to a lack of interpretation in the model.  
%One distinction observed. This set of differences is not identified by fairness measures because they are not associated with any of the biases we examined. Furthermore, we believe these cases are rare, making them difficult to identify through statistical measures. However, this qualitative analysis clearly shows that using a model causes variations in the assessments, which should be considered for technology adoption. 
%For example, 
\textcolor{update}{
% Table \ref{app:tab:error}
The 1\textsuperscript{st} row displays an essay with several misspellings (e.g., 
\textit{mont sanc} instead of \textit{mon sac}, and
\textit{cate d'idantite}  instead of \textit{carte d'identité}).
Indeed, these characteristics are typical and expected of A1-level candidates, but this profile does not reflect the real CEFR level of this candidate. 
The 2\textsuperscript{nd} row illustrates a case where a systematic spelling pattern (substituting \textit{e} with \textit{è}) yields a text easily understood by a native speaker but challenging for the system.
%shows an example that can be easily decoded by a native speaker, but is difficult for the system. In this case, the candidate consistently replaces the letter \textit{e} with \textit{è}, thus creating a text that is extremely different from any expected text.
This may have been the result, for example, of a keyboard usability issue or even an eye condition that was not previously identified during exam registration. 
% Regardless of the cause, both human raters decide that this replacement does not reflect the candidate's proficiency level when assigning an intermediate score.
% cela peut être un candidat avec une maladie cocculaire qui ne nous a rien signalé). Dans ce cas, notre modèle a complètement échoué. Fidelia attribue le niveau A1, alors que les correcteurs attribuent B1 et B2. De leur côté, les correcteurs ont choisi de ne pas sanctionner ce qui semble être une mauvaise utilisation du clavier ou un candidat avec des besoins spécifiques. Il me semble intéressant de rapporter ce cas.
%
%
This shows the value of an evaluation that combines different criteria, as well as the benefit of not relying on a single automated evaluation system.
}

%\textcolor{blue}{%In addition, we sought to examine more closely how heterogeneity in TCF test-taker backgrounds bears on fairness. Our quantitative fairness analysis—consistent with common practice—conditioned on language (and, indirectly, geographic origin) and task; however, the qualitative analysis indicated that additional background-related factors may also shape outcomes. 
%As a proxy for a more encompassing approach of test-taker background, we stratified model performance by TCF variant (e.g., general audience, citizenship/residence, Quebec immigration, etc.). 
%\textbf{TO DO ! } }

%Hétérogénéité (analyse par déclinaisons) : 
%	- performances plus faibles de ANF et CNF versus. TP/QC/DAP : cela s’explique par des profils plus hétérogènes ou nombres de données dans l’entraînement
%	- Est-ce que le modèle prendrait en compte les sujets indirectement, ce qui expliquerait la distribution particulière ?
%	- Certains niveaux sont constitués majoritairement de textes issus de certains déclinaisons

\subsection{Failure Identification Analysis}\label{sec:FailureIdentification}

% We used the Expected Calibration Error (ECE) \cite{guo2017calibration} to analyze the relationship between the models' accuracy and confidence. 
% This involves comparing the distribution of model predictions at each proficiency level to the expected essay levels. We treated the distribution of human rater scores as a model since human raters also exhibit some degree of error. In this analysis, we focus on the vote models. 
To explore the relationship between the models' accuracy and confidence with Expected Calibration Error, we focus on the \textit{vote} models by comparing the distribution of model predictions at each proficiency level to the expected essay levels.
%Similarly to human rater, we used the pool of models' predictions as the distribution for the vote model. Thus, each of the 10 models used to vote is considered as an independent rater. % Therefore, the probability of a specific score level is determined by the percentage of these 10 models that predict that score.
%
We observed the following measurement errors:
0.19 (0.001 of standard deviation) for human raters
0.15 (0.002) for \textit{SL\textsubscript{vote}}, and 
0.08 (0.004) for \textit{SC\textsubscript{vote}}, \textit{IV\textsubscript{vote}} and \textit{TR\textsubscript{vote}}.
These results indicate that the models are closer to the gold standard than the human raters. However, this should not be interpreted as the models achieving superhuman performance.
%On the contrary, it shows that models are capable of potentiating the biases that exist in human evaluation. 

% \textcolor{blue}{\st{Precisely in order to mitigate the biases of human assessment, certification organizations, such as FEI, adopts approaches such as different raters per essay and the grading of different essays per candidate.}}

%To further analyze the realiability of each of our 4 top architectures from each family, we checked the relation between the distribution of the predictions of the 10 models resulting from the cross-validation procedure and the annotators percentage agreement. 

Taking a deeper look at failure identification, we inspected a possible association between the distribution of the predictions of the 10 models and the raters percentage of agreement. 
%
%This analysis split the data into three groups: cases for which all our models are \textit{correct}, are all \textit{wrong}, or disagree (\textit{mixed}).\footnote{The three groups are created based on the prediction of all models to minimize possible random variations, as the models have similar performance.} 

When considering the average percentage of exact agreement between raters, we observed large differences between cases for which all four models are \textit{correct} (78\% of exact agreement), are \textit{wrong} (63\%), and 66\% for the \textit{mixed} group.
% is between them with average 67\% agreement when the model is correct and 65\% when it is wrong. 
Distinguishing correct and wrong cases within the \textit{mixed} set, we noticed that \textit{TR}, \textit{SC} and \textit{SL} tend to be correct when the raters have a higher agreement (68\% for mixed$_{correct}$ v. 64\% for mixed$_{wrong}$) while \textit{IV} shows the opposite trend (64\% for mixed$_{correct}$ v. 68\% for mixed$_{wrong}$).\footnote{Average agreement percentage by the 3 groups and the model correctness is shown in Table~\ref{ref:app:annotatorsAgreement} at Appendix \ref{app:feat_selection}.}
%This indicated an average difference of 5\% in the extreme cases (average agreement of 68\% when all models are correct and 63\% when they are all wrong). 

It is also worth noting that 94\% and 91\% of essays in the \textit{wrong} and the \textit{correct} groups respectively were rated by the small rater pool (9 raters), while 87\% of essays in the \textit{mixed} group were rated by 55 raters. %the small pool. 
This shows a tendency of models to disagree on the same cases as human raters.

\section{Discussion} \label{sec:disc}

Our results show that the systems we investigated are valid and competitive with humans. Nonetheless, they must be understood in the context of a broader evaluation framework: the discussion of our findings is therefore structured according to the five areas of \citet{williamson2012framework}, using the mapping proposed in Section~\ref{sec:Methodology}.

The results show minimal differences in correlations between human and system raters (see Section \ref{sec:annotation}), confirming a good \textit{(1) construct relevance and representation}. This indicates that both types of raters %exhibit \textbf{high agreement} and consistent scoring, indicating they 
similarly interpret and apply the rubric based on linguistic skills. The small average difference reflects this consistency. The slightly better alignment of models with linguistic features suggests humans might rely on criteria different from those encoded in the models. Moreover, the differences in correlation for cases of weak agreement might indicate a complementarity between human and automatic ratings.
% \textcolor{cyan}{Finally, these results discuss the linguistic features explored in this study. However, 
% %as pointed out in \textbf{Section \ref{res:comp}/\ref{res:comp:generalization}}, 
% AES literature lacks features addressing the advanced levels, particularly C2, which may }
% more features for capturing more linguistic aspects of the construct, and in this case humans raters seem to use different criteria.

For \textit{(2) association with typical scoring method (human scores)}, performance across architectures, including improvements from voting, demonstrates robustness from AES in ranking and identifying the essay levels.  
Automated systems outperform human raters across all metrics, indicating a potential for consistent and accurate assessments. 
%
%Concerning \textit{(3) association with independent measures}, 
The results of linguistic correlation in the samples with varying agreement levels illustrate the impact of human agreement on the scoring criteria. High consistency in full and close agreement samples suggests precise rubric adherence, while on challenging essays (weak agreement sample), the models remain reliant on linguistic features, whereas it seems that human raters also use other information.
This points out a likely improvement when combining human and automatic raters.
%High consistency between human and model raters points to a validates the criteria. Models show slightly better alignment with linguistic features, suggesting greater precision in capturing rubric constructs. Models' better alignment with linguistic features suggests enhanced precision. However, differences remain consistent across agreement levels.
\textcolor{update}{Despite these promising results, we must be cautious in interpreting them. Our analysis suggests a relationship between model errors and the amount of lexical errors made by candidates in an essay. %This raises future research questions about the completeness of AES models in relation to the expected task (i.e., are the models only detecting faults, or are they also detecting signs of linguistic competence?).
}

Regarding \textit{(4) generalizability of scores}, AES models demonstrate stable generalization. While human raters show a performance drop between the core and generalization corpora, AES models improve slightly. Human performance shifted due to the higher number of human raters involved in creating the generalization corpus, thus obtaining a more reliable scoring (in the core corpus, each rater typically contributes 50\% of the score decision, while in the generalization corpus a rater may impact up to 10\% of the final score). \textcolor{update}{In addition, we notice that the models' performances are underestimated in the core corpus. This is due to the limited number of raters (i.e., only two) %used to score an essay, 
which our findings show is insufficient. This goes in the same direction as studies in the language assessment field \cite{bouwer2015effect,huang2023rater} and raises questions about the performance estimation of several AES models trained on corpora where a single rater assigned the score.
In our study, this impact is most noticeable in the C2 level, which challenges the human raters.
Such findings spotlight the need for broader use of well-calibrated generalization corpora in the field.
%Following the current corpus design approach, this indicates that the creation of an accurate corpus would require various raters per essay (about 10 in this work), making the cost unfeasible.
}
Although this comes at a high processing cost, the \textit{vote} results show it is possible to improve the model's performance through ensemble methods and item response theory.

Regarding \textit{(5) score use and consequences}, 
the fairness analysis
on the core corpus reveals no language, gender or task bias, suggesting models generalize well under strong rater agreement. Biases observed on weak agreement samples highlight areas for improvement. The absence of bias on the generalization corpus demonstrates that the models can handle different populations fairly.
Moreover, the ECE study shows models are well-calibrated with fewer errors than human raters, reinforcing their reliability in high-stakes environments. Variability in human-rater agreement for correct and incorrect predictions emphasizes the need for cautious consideration. The varying trends in model performance based on rater agreement %(e.g., \textit{IV} model) 
suggest some models might be better suited to specific contexts or types of essays.

\section{Final Remarks}
\label{sec:ccl}

The main objective of our work was to address AES in the context of high-stakes language tests, specifically for FFL. This was motivated by the gap between existing AES evaluation standards and the high-stakes requirements of language certification.
Our study focused on assessing the validity of AES models, emphasizing construct validity as well as raters' agreement and model generalizability.
We systematically compared different AES modeling approaches using various evaluation metrics on two  corpora (core and generalization) for a better study of generalizability.
%We systematically compared different AES modeling approaches using various evaluation metrics across two degrees of result generalization.
We analyzed AES models' predictions in terms of alignment with the gold standard, the indirect relation to linguistic features associated with language proficiency, fairness of treatment for test-takers, and the monitoring of failures in a high-stake environment.

%(3) warnings on the pitfalls of AES models in an authentic language certification context; and
Our results are supported by substantial resources. The core corpus is the largest calibrated corpus examined in French AES for certification and the generalization corpus benefits from numerous human raters, supporting reliable evaluation estimates. We found that metrics that overlook exact predicted levels (e.g. QWK and correlation) can give an appearance of high performance when models fail to predict exact levels. On the other hand, metrics focusing solely on exact level prediction (e.g. F1 and exact accuracy) fail to capture whether models understand the level scale. Additionally, model fairness must be evaluated beyond the entire corpus, including on cases where humans disagree.

%(1) a AES evaluation framework for language certification based on the dual perspectives of agreement and generalisation, and construct validity;
We extended the standard AES ABV framework by operationalizing different abstract assessment areas with quantitative metrics, based on agreement, generalization, and construct validity. This bridges the gap between work on AES architectures and broader language assessment studies.
%(4) a new state-of-the-art model for French as a Foreign Language (FFL).
Our study also advances the state of the art in French AES by proposing new models that outperform existing approaches. To the best of our knowledge, we also conducted the largest corpus study of linguistic features for FFL to date, providing valuable data for machine learning models and informing studies on teaching and assessing FFL.

Our study focused on a single language, training corpus, and language model. Nevertheless, the concatenation and baseline architectures align with those explored for English. In particular, the results for \textit{SL} are encouraging and motivate further exploration with other languages. The limitation to one language model is due to the very limited number of models available for French, CamemBERT being the most widely used transformer encoder for French. Given the lack of comparative studies in French AES, we took a first step to fill this gap. Future research should compare architectures across diverse corpora for broader generalization. % We also highlight the large size of our corpus, which makes it ideal for deep learning models. The observed ranking may vary with corpus size.

%In conclusion, we identified a performance gap between human and automated scoring. 
In conclusion, our results are encouraging for applying AES models to high-stakes language certification tests in French. They also suggest that human raters consider elements beyond linguistic features captured in our AES models, stressing the complementarity between human and automated raters. FEI started using an automated scoring engine in conjunction with human raters. This is needed to closely and continuously monitor the model's performances and compensate for its shortcomings. For example, humans can detect outliers (such as off-topic answers), which the model cannot do. Another important aspect to monitor which might affect the model's performance is the constant renewal of prompts made by FEI. %Finally, the feedback of human raters is important also in the event a candidate would contest the scores they obtained.
%human raters are important as they can provide feedback and explanations if needed. This is required in the event a candidate would contest the scores they obtained, which is their right.

For future work, we plan to study the characteristics of essays that all models fail to predict correctly. We also plan to investigate the generalizability to other tasks and languages as well as the impact of corpus size on model performance.

\section{Acknowledgements}

This research has been partially funded by the Fonds de la Recherche Scientifique de Belgique (F.R.S.-FNRS) under the grant MIS/PGY F.4518.21 and T.0080.23, and also by a research convention with France Éducation International. Computational resources have been provided by the Consortium des Équipements de Calcul Intensif (CÉCI), funded by the Fonds de la Recherche Scientifique de Belgique (F.R.S.-FNRS) under Grant No. 2.5020.11 and by the Walloon Region.

% Entries for the entire Anthology, followed by custom entries
% \bibliography{tacl2021}
\bibliography{anthology,custom}
\bibliographystyle{acl_natbib}

\onecolumn

\appendix

%\section{Appendix}
%\label{sec:appendix}

\section{Hyperparameter selection}
\label{app:hPrams}
%\paragraph{Hyperparameters}

We explored the following hyperparameters: learn rate: 0.0005, 0.005, 1.00E-05, 2.00E-05, 3.00E-05, 4E-05, 5E-05; optimisation algorithm: adam, sgd; and the use of gradient clipping.
The selected hyperparameter are presented in Table \ref{tab:hyperparameter}.

%\footnote{Gradient clipping is a technique used in deep learning to prevent exploding gradients during training.
%It involves setting a threshold value to limit the magnitude of the gradients during training, which can help prevent numerical instability and make it easier to find a good solution. This technique is beneficial when training deep neural networks with many layers.: yes (value = 1), no
% ; and the use of gradient clipping.
%Since early stop is an exploited hyperparameter, the model should be trained until it reaches the maximum number of epochs.
%
% The two implicit models used the same hyperparameter range as the baseline transformers.

\begin{table}[h!]
\centering
\begin{tblr}{
  row{1} = {c},
  cell{2}{2} = {r},
  cell{2}{3} = {c},
  cell{2}{4} = {r},
  cell{3}{2} = {r},
  cell{3}{3} = {c},
  cell{3}{4} = {r},
  cell{4}{2} = {r},
  cell{4}{3} = {c},
  cell{4}{4} = {r},
  cell{5}{2} = {r},
  cell{5}{3} = {c},
  cell{5}{4} = {r},
  cell{6}{2} = {r},
  cell{6}{3} = {c},
  cell{6}{4} = {r},
  cell{7}{2} = {r},
  cell{7}{3} = {c},
  cell{7}{4} = {r},
  hlines,
  vlines,
}
\textbf{Model} & \textbf{learn rate} & \textbf{optimisation} & \textbf{clip} \\
TR             & 3E-05               & adam                  & no            \\
CM             & 2E-05               & adam                  & no            \\
C2             & 2E-05               & adam                  & no            \\
SC             & 2E-05               & adam                  & yes           \\
IF             & 2E-05               & adam                  & no            \\
IV             & 2E-05               & adam                  & no            
\end{tblr}
\caption{Selected hyperparameter for each model.}
\label{tab:hyperparameter}
\end{table}

\section{Core Corpus Size} \label{sec:app:corpussize}
The corpus used in our study consists of 27,683 essays categorized across 6 CEFR levels (A1 to C2). Each essay is associated with 3 different tasks, with the total distribution shown in Table \ref{tab:app:corpussize}. The corpus is representative of a wide range of language of use, and covering essays from different prompts.

% T1 = descriptive / T2 = Narrative T3 = argumentative
\begin{table}[h!]
\centering
\begin{tabular}{|l|r|r|r|r|r|r|} 
\cline{2-7}
\multicolumn{1}{l|}{} & \multicolumn{1}{l|}{Score} & \multicolumn{1}{l|}{Task: descriptive} & \multicolumn{1}{l|}{Task: narrative} & \multicolumn{1}{l|}{Task: argumentative} & \multicolumn{1}{l|}{\#prompt} & \multicolumn{1}{l|}{\#language}  \\ 
\hline
A1                    & 2883                              & 306                                  & 969                                  & 921                                  & 993                             & 82                                     \\ 
\hline
A2                    & 5678                              & 593                                  & 1761                                 & 1772                                 & 2145                            & 109                                    \\ 
\hline
B1                    & 5793                              & 694                                  & 1850                                 & 1883                                 & 2060                            & 126                                    \\ 
\hline
B2                    & 5881                              & 721                                  & 2012                                 & 1987                                 & 1882                            & 117                                    \\ 
\hline
C1                    & 5408                              & 651                                  & 1727                                 & 1966                                 & 1715                            & 104                                    \\ 
\hline
C2                    & 2040                              & 486                                  & 657                                  & 716                                  & 667                             & 54                                     \\ 
\hline
\textit{Total}        & 27683                             & 856                                  & 8976                                 & 9245                                 & 9462                            & 164                                    \\
\hline
\end{tabular}
\caption{Core corpus size}
\label{tab:app:corpussize}

\end{table}

\section{Model performance and agreement between raters}
Table \ref{ref:app:annotatorsAgreement} shows the average percentage agreement between raters, broken down by models performance (correct, mixed, wrong), models error (correct or wrong), and architecture.

\begin{table*}[h!]
\centering
\begin{tabular}{|l|l|r|r|r|r|} 
\hline
\multicolumn{1}{|c|}{\textbf{group}} & \multicolumn{1}{c|}{\textbf{case}} & \multicolumn{1}{c|}{\textbf{majority TR}} & \multicolumn{1}{c|}{\textbf{majority SC}} & \multicolumn{1}{c|}{\textbf{majority IV}} & \multicolumn{1}{c|}{\textbf{majority SL}}  \\ 
\hline
all right                     & corr                               & 78,17~\%                                   & 78,17~\%                                   & 78,17~\%                                   & 78,17~\%                                    \\ 
\hline
mixed                                & corr                               & 67,68~\%                                   & 67,68~\%                                   & 64,30~\%                                   & 67,53~\%                                    \\ 
\hline
mixed                                & wrong                              & 64,45~\%                                   & 64,45~\%                                   & 67,68~\%                                   & 64,47~\%                                    \\ 
\hline
all wrong                            & wrong                              & 62,81~\%                                   & 62,81~\%                                   & 62,81~\%                                   & 62,81~\%                                    \\
\hline
\end{tabular}
\caption{Average percentage agreement between raters for model correctness and error}
\label{ref:app:annotatorsAgreement}
\end{table*}

\section{Details of the selected feature} \label{app:feat_selection}

The 48 features were selected as portrayed in Table~\ref{tab:selected_features}. For each feature (column ``Variable''), we report its family, its correlation with the CEFR levels in the core corpus, and the motivation behind its integration into the final set.

% \begin{table}[]
%     \centering
%     \begin{tabular}{c|c|c}
%     \textbf{Group} & \textbf{Family} & \textbf{Feat.} \\
%     \hline
%     \textbf{Length based} & Word length & 1 \\
%      & Sentence length & 2 \\
%      \hline
%      \textbf{Lexical Features} & Lexical diversity & 4 \\
%       & Content Overlap & 1 \\
%       & Graded lexicon & 7 \\
%       & Lexical sophistication & 2 \\
%       & Lexical norms & 3 \\
%       & Orthographic neighbors & 3 \\
%       & Affixes & 1 \\
%     \hline
%     \textbf{Syntactic features} & Language development & 3 \\
%      & Dependency relations & 4 \\
%      & POS tag & 2 \\
%      & Morphology features & 3 \\
%      & Tense features & 3 \\
%      & Clause features & 1 \\
%     \hline
%     \textbf{Discursive features} & Text coherence & 1 \\
%      & Referential expressions & 2 \\
%      & Dialogue variables & 1 \\
%     \hline
%     \textbf{Errors features} & Grammatical errors & 1 \\
%      & Lexical errors & 1 \\
%      & Typographical errors & 1 \\
%      & All types of errors & 1 \\
%     \end{tabular}
%     \caption{Selected features}
%     \label{tab:selected_features}
% \end{table}

% \usepackage{hhline}

% \usepackage{multirow}
% \usepackage{hhline}

\begin{table*}[h!]
\centering
\resizebox{\textwidth}{!}{%
\begin{tabular}{|l|l|r|p{8.8cm}|} 
\hline
\multicolumn{1}{|c|}{\textbf{Family}} & \multicolumn{1}{c|}{\textbf{Variable}} & \multicolumn{1}{c|}{\textbf{Spearman}} & \multicolumn{1}{c|}{\textbf{Motivation}}                                                                      \\ 
\hline \hline
Word length                           & LENwrdSYL\_avg                         & 0,6                                    & Best Spearman, Pearson, mutual info gain across all levels                                                                \\
\hline \hline
Sentence Length                                     & LENsntWRD\_q1                          & 0,43                                   & Best Spearman across all levels                                                                                           \\
                                      & LENsntWRD\_q3                          & \textit{0,38}                          & Best mutual info gain + close to best Pearson                                                               \\ 
\hline \hline
Lexical diversity                       & LEXdvrWSR\_avg                         & 0,64                                   & Best Spearman, mutual info gain across all levels                                                              \\
                                      & LEXdvrFLR\_avg                         & \textit{0,61}                          & Best C2 vs all FLC\_avg                                                                                      \\
                                      & LEXdvrNSM\_avg                         & \textit{0,44}                          & Best B1 vs all (-0.07)                                                                                               \\
                                      & LEXdvrVSC\_avg                         & \textit{0,52}                          & Highly ranked across all families                                                                                                    \\ 
\hline \hline
Graded lexicons                       & LEXgrdFA1\_20P                         & -0,61                                  & Best Spearman across all levels                                                                                           \\
                                      & LEXgrdFA2\_20P                         & \textit{-0,56}                         & The levels between A2 and C2 have been added to provide                                 \\
                                      & LEXgrdFB1\_20P                         & \textit{-0,51}                         &        additional information for level A1.                                                                                                                   \\
                                      & LEXgrdFB2\_20P                         & \textit{-0,5}                          &                                                                                                                           \\
                                      & LEXgrdFC1\_20P                         & \textit{-0,39}                         &                                                                                                                           \\
                                      & LEXgrdFC2\_20P                         & \textit{-0,29}                         &                                                                                                                           \\
                                      & LEXgrdFMLA1\_rsd                       & \textit{0,52}                          & Best mutual info gain %+ B1 vs all (-0,15) + B2 vs C1 (-0,24)                                                         
                                      \\ 
\hline \hline
Orthographic neighbors                & LEXnghORT\_80P                         & 0,57                                   & Best Spearman, mutual info gain across all levers                                                              \\
                                      & LEXnghNUM\_20P                         & \textit{-0,55}                         & Best Pearson across all levels + A1 vs all, B2 vs all + A2 vs B1, B1 vs B2                \\
                                      & LEXnghPHO\_avg                         & \textit{0,52}                          & Best B1 vs all + B2 vs C1 (0,19)                                                                              \\ 
\hline \hline
Content Overlap                       & LEXcovLLST\_90P                        & 0,43                                   & Best Spearman, Person across all levels                                                                                \\ 
\hline \hline
Lexical sophistication                & LEXsopFK1\_avg                         & 0,47                                   & Best Spearman across all levels                                                                                          \\
                                      & LEXsopLLK3\_max                        & \textit{0,41}                          & Best A1 vs all + B2 vs all + different resource                                                         \\ 
\hline \hline
Lexical Norms                         & LEXnrmAOA\_max                         & 0,4                                    & Best Spearman and mutual info gain across all levels                          \\
                                      & LEXnrmIMG\_10P                         & \textit{-0,36}                         & Best Pearson across all levels + different resource                                                                     \\
                                      & LEXnrmFAM\_rsd                         & \textit{0,2}                           & Best B1 vs all (0,1) , B2 vs C1, C1 vs C2 + different resource                                       \\ 
\hline \hline
Morphology features                   & LEXafxA\_avg                           & \multicolumn{1}{r|}{0.48}                  & Information diversity % → Contrôler performance sur V2                                                                 
\\ 
\hline \hline
Language development                  & SYNdevHGT\_90P \_max                   & 0,52                                   & \_max Best Spearman, Pearson across all levels, 90P close                                                          
\\
                                      & SYNdevNPRSVPPART\_avg                  & \textit{0,31}                          & Best B1 vs all (0,08)                                                                                                \\
                                      & SYNdevSIMA\_90P                        & \textit{0,41}                          & Best C1 vs C2 (0,08)                                                                                                 \\ 
\hline \hline
Dependency Relations                  & SYNdepMARK\_90P                        & 0,47                                   & \_max Best Spearman across all levels, 90P close                                                                    \\
                                      & SYNdepREPARANDUM\_kurtosis             & \textit{-0,41}                         & Best Pearson across all levels                                                                                            \\
                                      & SYNdepCOP\_avg                         & \textit{0,04}                          & Best C1 vs C2                                                                                                \\
                                      & SYNdepACL\_dolch                       & \textit{0,37}                          & Best B1 vs all + \_max B1 vs B2, B2 vs C1                                                       \\ 
\hline \hline
POS Tag                               & SYNposADP\_avg                         & 0,31                                   & Best Spearman across all levels                                                                                           \\
                                      & SYNposSCONJ\_skewness                  & 0,31                                   & Best Pearson across all levels                                                                                            \\ 
\hline \hline
Clause features                       & SYNclsTYPCX\_90P                       & 0,46                                   & Best Spearman across all levels                                                                                           \\ 
\hline \hline
Morphology features                   & SYNmorVERBFORM\_FIN\_20P               & -0,38                                  & Best Spearman across all levels                                                                                           \\
                                      & SYNmorNUMBER\_PLUR\_var                & \textit{0,33}                          & Best A1 vs A2 + interetsing information                                                                             \\
                                      & SYNmorVERBFORM\_PART\_90P              & \textit{0,27}                          & Best C1 vs C2 + information on time                                                                       \\ 
\hline \hline
                                      
Tense features                        & SYNtnsfINDP\_avg                       & -0,42                                  & Best Spearman, Pearson, MutInfo across all levels                                                                         \\
                                      & SYNtnsfPP\_kurtosis                    & \textit{0,22}                          & Best B1 vs all + information on time                                                                      \\
                                      & SYNtnsfTAP\_skewness                   & \textit{0,3}                           & Best C1 vs C2 (0,05) + information on time                                                                       \\ 
\hline \hline
Referential expressions               & DISrefDN\_avg                          & 0,24                                   & Best Spearman, Pearson across all levels                                                                                  \\
                                      & DISrefPRSW\_avg                        & \textit{-0,19}                         & Best C1 vs all (0,11) + B1 vs B2 (-0,09)                                                                             \\ 
\hline \hline
Dialogue Variables                    & DISdiaPPEI2\_skewness                  & \textit{0,19}                          & Best Pearson, MutInfo across all levels + A2 vs B1, B1 vs B2                                                \\ 
\hline \hline

Text coherence                        & DIScohLSALADJ\_90P                     & 0,46                                   & Best Pearson, MutInfo across all levels + close for Spearman across all levels + best B2 vs C1, C1 vs C2  \\ 
\hline \hline
Grammatical errors                    & ERRgrmSUM\_avg                         & -0,73                                  & Best Spearman, Pearson, MutInfo across all levels                                                                         \\
                                      & ERRlexSUM\_avg                         & -0,32                                  & Best Spearman , Pearson MutInfo across all levels                                                                         \\
                                      & ERRtypSUM\_avg                         & -0,69                                  & Best Spearman, Pearson, MutInfo across all levels                                                                         \\
                                      & ERRallSUM\_avg                         & -0,79                                  & Best Spearman, Pearson, MutInfo across all levels                                                                         \\
\hline
\end{tabular}
}
    \caption{48 selected features. For an explanation about them, see \url{https://cental.uclouvain.be/fabra/docs_expert.html}.}
    \label{tab:selected_features}
\end{table*}

\clearpage
\section{Results of the direct comparison and linguistic correlation}

\begin{table*}[!h]
\centering
\begin{tabular}{|rl|c|r|r|r|r|r|r|r|r|} 
\hline
\multicolumn{2}{|l|}{\multirow{2}{*}{\textbf{Corpus}}}                                        & \multirow{2}{*}{\textbf{Model}} & \multicolumn{7}{c|}{\textbf{Direct Comparison}}                                                                                                                              & \multicolumn{1}{c|}{\multirow{2}{*}{\begin{tabular}[c]{@{}c@{}}\textbf{$\Delta$}\\\textbf{Corr}\end{tabular}}}  \\
\multicolumn{2}{|l|}{}                                                                        &                                 & \multicolumn{1}{r}{$\rho$} & \multicolumn{1}{r}{$\kappa$} & \multicolumn{1}{r}{$QWK$} & \multicolumn{1}{r}{$MSE$} & \multicolumn{1}{r}{$EA$} & \multicolumn{1}{r}{$AA$} & $F1$               & \multicolumn{1}{c|}{}                                                                                    \\ 
\hline
\multicolumn{2}{|l|}{\multirow{9}{*}{\begin{tabular}[c]{@{}l@{}}Core\\Corpus\end{tabular}}}   & TR                              & 0.90                    & 0.48                      & 0.89                    & 0.50                    & 0.57                   & 0.98                   & 0.57             & -0.02                                                                                                    \\ 
\cline{3-11}
\multicolumn{2}{|l|}{}                                                                        & CM                              & 0.90                    & 0.48                      & 0.89                    & 0.49                    & 0.57                   & 0.98                   & 0.57             & -0.02                                                                                                    \\ 
\cdashline{3-11}
\multicolumn{2}{|l|}{}                                                                        & C2                              & 0.90                    & 0.48                      & 0.89                    & 0.49                    & 0.57                   & 0.98                   & 0.57             & -0.02                                                                                                    \\ 
\cdashline{3-11}
\multicolumn{2}{|l|}{}                                                                        & SC                              & 0.90                    & 0.44                      & 0.88                    & 0.55                    & 0.54                   & 0.97                   & 0.54             & -0.02                                                                                                    \\ 
\cline{3-11}
\multicolumn{2}{|l|}{}                                                                        & IF                              & 0.90                    & 0.48                      & 0.89                    & 0.48                    & 0.58                   & 0.98                   & 0.57             & -0.02                                                                                                    \\ 
\cdashline{3-11}
\multicolumn{2}{|l|}{}                                                                        & IV                              & 0.90                    & 0.48                      & 0.89                    & 0.49                    & 0.57                   & 0.98                   & 0.57             & -0.01                                                                                                    \\ 
\cline{3-11}
\multicolumn{2}{|l|}{}                                                                        & SL                              & 0.90                    & 0.50                      & 0.89                    & 0.44                    & 0.60                   & 0.99                   & 0.59             & -0.03                                                                                                    \\ 
\cdashline{3-11}
\multicolumn{2}{|l|}{}                                                                        & HL                              & 0.90                    & 0.47                      & 0.88                    & 0.48                    & 0.57                   & 0.98                   & 0.56             & -0.02                                                                                                    \\ 
\cline{3-11}
\multicolumn{2}{|l|}{}                                                                        & Raters                          & 0.91                    & 0.70                      & 0.91                    & 0.40                    & 0.75                   & 0.98                   & 0.75             & 0.01                                                                                                     \\ 
\hline \hline
\vcell{}         & \multirow{9}{*}{\begin{tabular}[c]{@{}l@{}}Full\\agreement\end{tabular}}           & \vcell{TR}                      & \vcell{0.92}            & \vcell{0.55}              & \vcell{0.91}            & \vcell{0.40}            & \vcell{0.63}           & \vcell{0.99}           & \vcell{0.64}     & \vcell{-0.01}                                                                                            \\[-\rowheight]
\printcellmiddle &                                                                            & \printcellmiddle                & \printcellmiddle        & \printcellmiddle          & \printcellmiddle        & \printcellmiddle        & \printcellmiddle       & \printcellmiddle       & \printcellmiddle & \printcellmiddle                                                                                         \\ 
\cline{3-11}
                 &                                                                            & CM                              & 0.92                    & 0.56                      & 0.91                    & 0.39                    & 0.64                   & 0.99                   & 0.64             & -0.01                                                                                                    \\ 
\cdashline{3-11}
                 &                                                                            & C2                              & 0.92                    & 0.55                      & 0.91                    & 0.40                    & 0.63                   & 0.99                   & 0.63             & -0.01                                                                                                    \\ 
\cdashline{3-11}
                 &                                                                            & SC                              & 0.92                    & 0.51                      & 0.90                    & 0.45                    & 0.60                   & 0.98                   & 0.60             & -0.01                                                                                                    \\ 
\cline{3-11}
                 &                                                                            & IF                              & 0.92                    & 0.56                      & 0.91                    & 0.38                    & 0.64                   & 0.99                   & 0.64             & -0.01                                                                                                    \\ 
\cdashline{3-11}
                 &                                                                            & IV                              & 0.92                    & 0.55                      & 0.91                    & 0.39                    & 0.63                   & 0.99                   & 0.64             & -0.01                                                                                                    \\ 
\cline{3-11}
                 &                                                                            & SL                              & 0.92                    & 0.57                      & 0.91                    & 0.36                    & 0.66                   & 0.99                   & 0.65             & -0.02                                                                                                    \\ 
\cdashline{3-11}
                 &                                                                            & HL                              & 0.91                    & 0.56                      & 0.91                    & 0.38                    & 0.64                   & 0.99                   & 0.63             & -0.01                                                                                                    \\ 
\cline{3-11}
                 &                                                                            & Raters                          & 1.00                    & 1.00                      & 1.00                    & 0.00                    & 1.00                   & 1.00                   & 1.00             & -0.00                                                                                                    \\ 
\hline \hline
\vcell{}         & \multirow{9}{*}{\begin{tabular}[c]{@{}l@{}}Close\\agreement\end{tabular}}           & \vcell{TR}                      & \vcell{0.89}            & \vcell{0.41}              & \vcell{0.87}            & \vcell{0.59}            & \vcell{0.52}           & \vcell{0.97}           & \vcell{0.51}     & \vcell{-0.02}                                                                                            \\[-\rowheight]
\printcellmiddle &                                                                            & \printcellmiddle                & \printcellmiddle        & \printcellmiddle          & \printcellmiddle        & \printcellmiddle        & \printcellmiddle       & \printcellmiddle       & \printcellmiddle & \printcellmiddle                                                                                         \\ 
\cline{3-11}
                 &                                                                            & CM                              & 0.89                    & 0.40                      & 0.87                    & 0.59                    & 0.51                   & 0.97                   & 0.50             & -0.02                                                                                                    \\ 
\cdashline{3-11}
                 &                                                                            & C2                              & 0.89                    & 0.42                      & 0.87                    & 0.56                    & 0.53                   & 0.97                   & 0.51             & -0.02                                                                                                    \\ 
\cdashline{3-11}
                 &                                                                            & SC                              & 0.89                    & 0.37                      & 0.86                    & 0.65                    & 0.48                   & 0.96                   & 0.47             & -0.02                                                                                                    \\ 
\cline{3-11}
                 &                                                                            & IF                              & 0.89                    & 0.41                      & 0.87                    & 0.57                    & 0.52                   & 0.97                   & 0.50             & -0.02                                                                                                    \\ 
\cdashline{3-11}
                 &                                                                            & IV                              & 0.89                    & 0.41                      & 0.87                    & 0.57                    & 0.52                   & 0.97                   & 0.50             & -0.02                                                                                                    \\ 
\cline{3-11}
                 &                                                                            & HL                              & 0.89                    & 0.45                      & 0.88                    & 0.50                    & 0.55                   & 0.98                   & 0.53             & -0.03                                                                                                    \\ 
\cdashline{3-11}
                 &                                                                            & SL                              & 0.88                    & 0.40                      & 0.86                    & 0.56                    & 0.52                   & 0.98                   & 0.49             & -0.02                                                                                                    \\ 
\cline{3-11}
                 &                                                                            & Raters                          & 0.87                    & 0.40                      & 0.86                    & 0.61                    & 0.51                   & 0.99                   & 0.50             & 0.01                                                                                                     \\ 
\hline \hline
\vcell{}         & \multirow{9}{*}{\begin{tabular}[c]{@{}l@{}}Weak\\agreement\end{tabular}}           & \vcell{TR}                      & \vcell{0.84}            & \vcell{0.23}              & \vcell{0.82}            & \vcell{0.87}            & \vcell{0.36}           & \vcell{0.92}           & \vcell{0.36}     & \vcell{-0.05}                                                                                            \\[-\rowheight]
\printcellmiddle &                                                                            & \printcellmiddle                & \printcellmiddle        & \printcellmiddle          & \printcellmiddle        & \printcellmiddle        & \printcellmiddle       & \printcellmiddle       & \printcellmiddle & \printcellmiddle                                                                                         \\ 
\cline{3-11}
                 &                                                                            & CM                              & 0.84                    & 0.21                      & 0.82                    & 0.86                    & 0.34                   & 0.94                   & 0.35             & -0.05                                                                                                    \\ 
\cdashline{3-11}
                 &                                                                            & C2                              & 0.83                    & 0.24                      & 0.82                    & 0.86                    & 0.37                   & 0.93                   & 0.36             & -0.05                                                                                                    \\ 
\cdashline{3-11}
                 &                                                                            & SC                              & 0.83                    & 0.24                      & 0.82                    & 0.91                    & 0.37                   & 0.91                   & 0.37             & -0.06                                                                                                    \\ 
\cline{3-11}
                 &                                                                            & IF                              & 0.83                    & 0.23                      & 0.81                    & 0.85                    & 0.36                   & 0.93                   & 0.34             & -0.05                                                                                                    \\ 
\cdashline{3-11}
                 &                                                                            & IV                              & 0.83                    & 0.24                      & 0.82                    & 0.85                    & 0.37                   & 0.93                   & 0.35             & -0.05                                                                                                    \\ 
\cline{3-11}
                 &                                                                            & SL                              & 0.83                    & 0.21                      & 0.81                    & 0.84                    & 0.35                   & 0.94                   & 0.33             & -0.06                                                                                                    \\ 
\cdashline{3-11}
                 &                                                                            & HL                              & 0.82                    & 0.18                      & 0.80                    & 0.88                    & 0.32                   & 0.94                   & 0.30             & -0.06                                                                                                    \\ 
\cline{3-11}
                 &                                                                            & Raters                          & 0.67                    & 0.33                      & 0.66                    & 1.88                    & 0.45                   & 0.57                   & 0.44             & 0.06                                                                                                     \\
\hline
\end{tabular}
    \caption{Results of the direct comparison and linguistic correlation analysis in the core corpus, including its splits by human raters agreement.}
    \label{tab:resultsFull}
\end{table*}

\clearpage
\section{Correlations Between Candidates' Errors and Model Residuals}
\label{app:corr_errors}

% \begin{table}[h]
%     \centering
%     \begin{tabular}{|c|S|S|S|S|}
%     \hline
%     \textbf{CEFR} & \textbf{\#docs} & \textbf{Lexical} & \textbf{Morphological} & \textbf{Syntactic} \\
%     A1 & 28 & 0.35$^{**}$ &0.15 & 0.14$ \\
%         A2 & 61 & 0.25$^{**}$ & 0.14 & 0.14$ \\
%         B1 & 66 & 0.14 & 0.08& 0.08$ \\
%         B2 & 68 & 0.08 & 0.04 & 0.05$ \\
%         C1 & 61 & 0.05 & 0.02 & 0.03$ \\
%         C2 & 24 & 0.03$^{**}$ & 0.01 & 0.02$^{*}$ \\ \hline
%     \hline
%         A1 & 28 & -0.53$^{**}$ &-0.08 & -0.28$ \\
%         A2 & 61 & -0.45$^{**}$ & -0.19 & -0.20$ \\
%         B1 & 66 & -0.03 & -0.03& -0.10$ \\
%         B2 & 68 & -0.07 & -0.03 & -0.05$ \\
%         C1 & 61 & 0.09 & 0.04 & 0.20$ \\
%         C2 & 24 & 0.62$^{**}$ & 0.08 & 0.46$^{*}$ \\ \hline
        
%     \end{tabular}
%     \caption{Correlations between presence of candidates' errors in the texts and the \textit{soft-labeling} model's residuals, by CEFR level and error type. Correlations with p-value $\leq$ 0.05 are marked with $^\star$ and $\leq$ 0.01 are marked with $^{\star\star}$.}
%     \label{tab:corr_errors}
    
% \end{table}

\begin{table}[!h]
    \centering
    \begin{subtable}[t]{\textwidth}
        \centering
        \begin{tabular}{|c|S|S|S|S|}
        \hline
        \textbf{CEFR} & \textbf{\#docs} & \textbf{Lexical} & \textbf{Morphological} & \textbf{Syntactic} \\
        \hline
        A1 & 28 & 0.35$^{**}$ &0.15 & 0.14 \\
        A2 & 61 & 0.25$^{**}$ & 0.14 & 0.14 \\
        B1 & 66 & 0.14 & 0.08& 0.08 \\
        B2 & 68 & 0.08 & 0.04 & 0.05 \\
        C1 & 61 & 0.05 & 0.02 & 0.03 \\
        C2 & 24 & 0.03$^{**}$ & 0.01 & 0.02$^{*}$ \\
        \hline
        \end{tabular}
        \caption{Proportion of errors per essay}
        \label{tab:corr_errors:proportion}
    \end{subtable}

    \vspace{0.5em} % espaço entre as tabelas (ajuste se quiser)

    \begin{subtable}[t]{\textwidth}
        \centering
        \begin{tabular}{|c|S|S|S|S|}
        \hline
        \textbf{CEFR} & \textbf{\#docs} & \textbf{Lexical} & \textbf{Morphological} & \textbf{Syntactic} \\
        \hline
        A1 & 28 & -0.53$^{**}$ &-0.08 & -0.28 \\
        A2 & 61 & -0.45$^{**}$ & -0.19 & -0.20 \\
        B1 & 66 & -0.03 & -0.03& -0.10 \\
        B2 & 68 & -0.07 & -0.03 & -0.05 \\
        C1 & 61 & 0.09 & 0.04 & 0.20 \\
        C2 & 24 & 0.62$^{**}$ & 0.08 & 0.46$^{*}$ \\
        \hline
        \end{tabular}
        \caption{Correlations between presence of candidates' errors in the texts}
        \label{tab:corr_errors:corr}
    \end{subtable}

    \caption{Error annotation in the \textit{soft-labeling} model's residuals, by CEFR level and error type. Correlations with p-value $\leq$ 0.05 are marked with $^\star$ and $\leq$ 0.01 are marked with $^{\star\star}$.}
    \label{tab:corr_errors}
\end{table}

\clearpage
\section{Examples of essays that challenge the model} \label{app:model_errors}

\begin{table}[!h]
\centering
\begin{tabular}{|c|c|p{5cm}|p{5cm}|} 
\hline
\multicolumn{2}{|c|}{\textbf{Level}} & \multicolumn{2}{c|}{\textbf{Essay}}                                                                                                                                                                                                                                                                                                                                                                                                                                                                                                                                                                                                                                                                                                                                                                                                                                                                                                                                                                                                                                                                                                                                                                                            \\
\textbf{Model} & \textbf{Humans}     & \multicolumn{1}{c|}{\textbf{Original}}                                                                                                                                                                                                                                                                                                                                                                                                                                                                                                                                                                                             & \multicolumn{1}{c|}{\textbf{English Translated}}                                                                                                                                                                                                                                                                                                                                                                                                                                                                                                                                         \\ 
\hline
A1             & B1                  & bonjour nicolas comme va tu . de mon cote sa ne va pas deus voles~ sont entre chez moi et il ont vole mont sanc il~ ya~ tout~ mes papiers~ j'ai porte pleinte a la gandameri~~ menteman j'ai plus de passeport ni de cate d'idandite bonne journne a toi et je te tien ou courent si la polise                                                                                                                                                                                                                                                                                                                                     & hello nicolas, how are you. as for me things are not going well two robbers came into my house and stole my backpack there is all my papers I filed a complaint with the police I do not have a passport or ID card have a good day and I'll let you know if the police.                                                                                                                                                                                                                                                                                                \\ 
\hline
A1             & B1/B2               & bonjours chèrès intèrnautès jèspèrs què vous allèr bièn. jai ètè invitè lè wèènkènd passè lors du mariagè organisè par la famillè . jè suis hypèr èxcistè je vous fairè part dè cèttè cèromoniè car cè qui ma intèrrèsè dunè part il yavait lès joliè dècorations ,idèniablè , dè la boissons dè lambiancès jusqua laubè la mariè ètait bèllè oh lalal ainsi què celui du mariè.ènsuitè ,cè mariagè ètait spèstaculairè parmis tans dè cèrèmoniè què jai ètait invitè.pour cèla je vous invitè a assitèr lès cèrèmoniès qui sèront important pou lè bièn ètrè dè votrè èpanouissèmènt jè vouys rèmèrciè a bièntot. & hello dear internet users, I hope you are well. I was invited last weekend to the wedding organized by the family. I'm super excited, I'm sharing this ceremony with you because what interested me on the one hand there were the pretty decorations, idèniable, from the party drinks until dawn the bride was beautiful oh lalal as well as that of the bridegroom. Then, this wedding was spectacular among so many ceremonies that I was invited. For this I invite you to attend the ceremonies which will be important for your well-being.I thank you see you soon.  \\
\hline
\end{tabular}
\caption{Examples of challenging essays}
\label{app:tab:error}
\end{table}

\clearpage
\section{Fine-grained inspection of the results}

\begin{table}[h]
\centering
\textcolor{update}{
\resizebox{.49\textwidth}{!}{
\begin{tabular}{|l|l|l|l|l|l|l|} 
\cline{2-7}
\multicolumn{1}{l|}{} & \multicolumn{1}{c|}{\textbf{A1}} & \multicolumn{1}{c|}{\textbf{A2}} & \multicolumn{1}{c|}{\textbf{B1}} & \multicolumn{1}{c|}{\textbf{B2}} & \multicolumn{1}{c|}{\textbf{C1}} & \multicolumn{1}{c|}{\textbf{C2}}  \\ \hline
TR                    &  0.66                 &  0.62                 &  0.57                 &  0.51                 &  0.55                 &  0.54                  \\ \hline
CM                    &  0.62                 &  0.62                 &  0.57                 &  0.53                 &  0.54                 &  0.54                  \\ \hdashline
C2                    &  0.65                 &  0.63                 &  0.55                 &  0.54                 &  0.56                 &  0.52                  \\ \hdashline
SC                    &  0.62                 &  0.61                 &  0.54                 &  0.49                 &  0.48                 &  0.48                  \\ \hline
IF                    &  0.62                 &  0.64                 &  0.56                 &  0.54                 &  0.56                 &  0.48                  \\ \hdashline
IV                    &  0.62                 &  0.63                 &  0.56                 &  0.52                 &  0.57                 &  0.53                  \\ \hline
SL                    &  0.65                 &  0.63                 &  0.58                 &  0.58                 &  0.62                 &  0.47                  \\ \hdashline
HL                    &  0.58                 &  0.60                 &  0.57                 &  0.55                 &  0.62                 &  0.44                  \\ \hline
Raters                & 0.76        & 0.76        & 0.74        & 0.75        & 0.76        & 0.74         \\ \hline
SL-GC           & 0.57        & 0.59        & 0.71        & 0.74        & 0.71        &  0.68         \\ \hline
\end{tabular}
}
\caption{F1 by CEFR level on the Core corpus and the SL model on the Generalization corpus (GC).}
\label{tab:f1by_level}
}
\end{table}

\end{document}